\documentclass[10pt,twocolumn,letterpaper]{article}

\usepackage{iccv}
\usepackage{times}
\usepackage{epsfig}
\usepackage{graphicx}
\usepackage{amsmath}
\usepackage{amssymb}
\usepackage{booktabs}
\usepackage{xcolor}
\usepackage{multirow}
\usepackage{arydshln}
\usepackage{threeparttable}
\usepackage{tabularx}
\usepackage{wrapfig}
\usepackage{url}
\usepackage{xspace}
\usepackage{subcaption}
\usepackage{enumitem}
\usepackage{algorithm}
\usepackage{algorithmic}
\usepackage{float}
\usepackage{wrapfig}
\usepackage{amsmath,amsfonts,bm}

\def\eqref#1{equation~\ref{#1}}

\def\1{\bm{1}}

\def\wrt{\mbox{w.r.t.}}
\def\etal{\emph{et al.}}

\def\mB{{\mathcal B}}

\def\mD{{\mathcal D}}

\def\mL{{\mathcal L}}

\def\mS{{\mathcal S}}

\def\0{{\bf 0}}
\def\1{{\bf 1}}

\def\bg{{\bf g}}

\def\bm{{\bf m}}

\def\bu{{\bf u}}

\def\bw{{\bf w}}
\def\bx{{\bf x}}

\usepackage[breaklinks=true,bookmarks=false]{hyperref}

\iccvfinalcopy

\definecolor{customred}{HTML}{FE0013}
\definecolor{customgreen}{HTML}{1caf58}
\definecolor{customblue}{HTML}{00b1ed}
\definecolor{customorange}{HTML}{ed7c3b}
\definecolor{custompurple}{HTML}{BF9BF3}

\definecolor{citecolor}{HTML}{0071bc}
\definecolor{paleplum}{rgb}{0.8, 0.6, 0.8}
\hypersetup{
  colorlinks,
  citecolor=citecolor,
  linkcolor=red
}

\ificcvfinal\fi

\newlength\savewidth
  
  \makeatletter
\def\@fnsymbol#1{\ensuremath{\ifcase#1\or \dagger\or \ddagger\or
   \mathsection\or \mathparagraph\or \|\or **\or \dagger\dagger
   \or \ddagger\ddagger \else\@ctrerr\fi}}
\makeatother

\begin{document}

\title{Sharpness-Aware Quantization for Deep Neural Networks}

\author{
Jing Liu
\quad Jianfei Cai
\quad Bohan Zhuang\thanks{Corresponding author. Email: $\tt bohan.zhuang@gmail.com$} \\[0.2cm]
ZIP Lab, Monash University
}

\maketitle
\ificcvfinal\thispagestyle{empty}\fi

\newcommand{\tabincell}[2]{\begin{tabular}{@{}#1@{}}#2\end{tabular}}

\newcommand{\methodfullname}{Sharpness-Aware Quantization \xspace}
\newcommand{\methodshortname}{SAQ\xspace}

\def\jing{\textcolor{black}}
\def\revise{\textcolor{black}}

\begin{abstract}
Network quantization is a dominant paradigm of model compression.
However, the abrupt changes in quantized weights during training often lead to 
severe loss fluctuations and result in a sharp loss landscape, making the gradients unstable and thus degrading the performance.
Recently, Sharpness-Aware Minimization (SAM) has been proposed to smooth the loss landscape and improve the generalization performance of the models.
Nevertheless, directly applying SAM to the quantized models can lead to perturbation mismatch or diminishment issues, resulting in suboptimal performance.
In this paper, we propose a novel method, dubbed Sharpness-Aware Quantization (SAQ), 
to explore the effect of SAM in model compression, particularly quantization for the first time. Specifically, we first provide a unified view of quantization and SAM by treating them as introducing quantization noises and adversarial perturbations to the model weights, respectively. According to whether the noise and perturbation terms depend on each other, SAQ can be formulated into three cases, which are analyzed and compared comprehensively. Furthermore, by introducing an efficient training strategy, SAQ only incurs a little additional training overhead compared with the default optimizer (e.g., SGD or AdamW).
Extensive experiments on both convolutional neural networks and Transformers across various datasets (i.e., ImageNet, CIFAR-10/100, Oxford Flowers-102, Oxford-IIIT Pets) show that SAQ improves the generalization performance of the quantized models, yielding the SOTA results in uniform quantization. For example, on ImageNet, SAQ outperforms AdamW
by 1.2\% on the Top-1 accuracy for 4-bit ViT-B/16. Our 4-bit ResNet-50 surpasses the previous SOTA method by 0.9\% on the Top-1 accuracy.
\end{abstract}

\section{Introduction}
\label{sec:introduction}
With powerful high-performance computing and massive labeled data, 
convolutional neural networks (CNNs) and Transformers
have dramatically improved the accuracy of many computer vision (CV) and natural language processing (NLP) tasks, such as image classification~\cite{he2016deep,dosovitskiy2020image}, dense prediction~\cite{ren2015faster,carion2020end}, sentence classification~\cite{wang2018glue,devlin2018bert},
and machine translation~\cite{McCann2017LearnedIT,vaswani2017attention},
to the level of being ready for real-world applications. 
Despite the remarkable breakthroughs that deep learning has achieved, the considerable computational overhead and model size greatly hampers the development and deployment of deep learning techniques at scale, especially on resource-constrained devices such as mobile phones.
To obtain compact models, many network quantization methods~\cite{hubara2016binarized,zhou2016dorefa} have been proposed to tackle the efficiency bottlenecks.

\begin{figure}[t]
    \begin{subfigure}[b]{0.215\textwidth}
         \centering
         \includegraphics[height=1.2in]{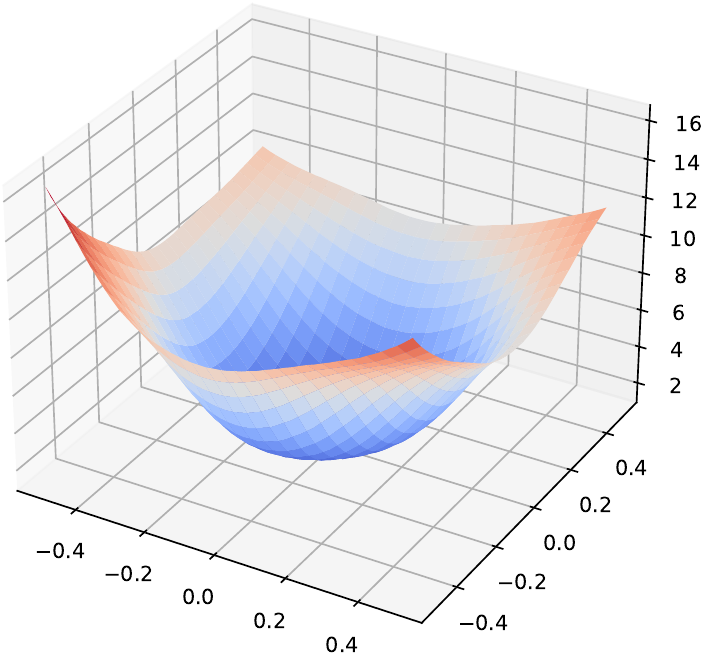}
         \caption{Full-precision ResNet-18}
         \label{fig:full_precision}
    \end{subfigure}
    \begin{subfigure}[b]{0.245\textwidth}
         \centering
         \includegraphics[height=1.2in]{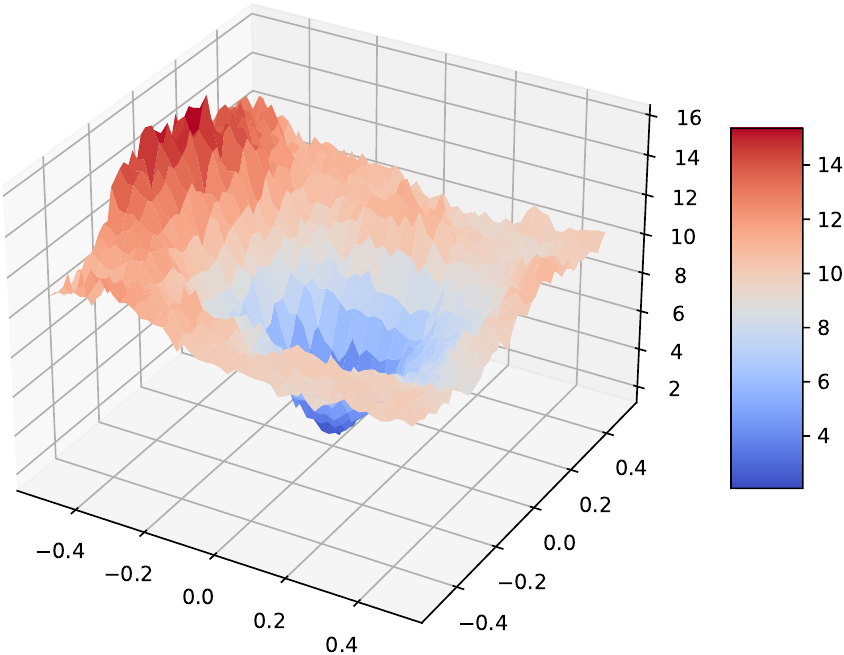}
         \caption{2-bit ResNet-18}
         \label{fig:low_precision}
    \end{subfigure}
    \vspace{-0.07in}
    \caption{The loss landscapes of the full-precision and 2-bit ResNet-18 \jing{models} on ImageNet. 
    We plot the loss landscapes using the visualization method in~\cite{Li2018VisualizingTL}. 
    More visualizations can be found in the supplementary.}
    \label{fig:loss_landscape}
    \vspace{-0.3in}
\end{figure}

Despite the high compression ratio, training a low-precision model is very challenging due to the discrete and non-differentiable nature of network quantization. In contrast to the full-precision ones, the low-precision models represent weights, activations, and even gradients with only a small set of values, which limits the representation power of the quantized models. As shown in Figure~\ref{fig:loss_landscape}, a slight change in full-precision weights coming from the gradient update or quantization noises might incur \jing{large} change in quantized weights due to discretization, which leads to \jing{drastic} loss fluctuations and results in much sharper loss landscape~\cite{liu2021adam}. As a result, the enormous loss fluctuations make gradients unreliable during optimization, which misleads weight update and thus incurs a performance drop. 

There have been some studies showing that flat minima of the loss function found by stochastic gradient-based methods result in good generalization~\cite{hochreiter1995simplifying,keskar2017large,dziugaite2017computing,jiang2019fantastic}. Recently, Sharpness-\jing{Aware} Minimization (SAM)~\cite{foret2021sharpnessaware} and its variants~\cite{Kwon2021ASAMAS,zhuang2021surrogate,kim2022fisher} have been proposed to smooth the loss landscape and significantly improve model generalization ability. Specifically, SAM first introduces perturbations to model weights and then minimizes a perturbed loss to seek parameters that lie in neighborhoods with uniformly low training loss.
However, all the existing methods are based on full-precision over-parameterized models. \textit{How to perform SAM on the compressed models, especially on the quantized ones, has rarely been explored,  which is a new and important problem.} \revise{A simple solution is directly applying SAM to train the quantized models. Nevertheless, as we will discuss in Section~\ref{sec:saq_different_cases}, the introduced perturbations can be either mismatched with the quantized weights or diminished by clipping and discretization operations, which may lead to suboptimal performance.}

In this paper, we propose a novel method, called \methodfullname (\methodshortname), to find minima with both low loss value and low loss curvature and thus improve the generalization performance of the quantized models. \textit{To our knowledge, this is a pioneering work to study the effect of SAM in model compression, especially in network quantization.} To this end, we first provide a unified view for quantization and SAM, where we %
treat
them as introducing quantization noises $\boldsymbol{\epsilon}_q$ and adversarial perturbations $\hat{\boldsymbol{\epsilon}}_s$ to the model weights, respectively. According to whether $\boldsymbol{\epsilon}_q$ and $\hat{\boldsymbol{\epsilon}}_s$ are dependent on each other, we can %
formulate
our SAQ into three cases. We then study and compare these cases comprehensively. Considering that SAQ requires additional training overhead to compute $\hat{\boldsymbol{\epsilon}}_s$, we further introduce an efficient training strategy, enabling SAQ to achieve comparable training efficiency %
as
the
default
optimization counterpart such as AdamW or SGD, which makes it scalable to large models. 
Extensive experiments on both CNNs and Transformers \revise{across various datasets} show the promising performance of \methodshortname.

Our main contributions are summarized as follows:
\begin{itemize}[leftmargin=*]
\item
We propose \methodshortname to seek flatter minima for the quantized models \jing{in order to} materially improve the generalization performance. To our knowledge, \jing{this is a} pioneering work \jing{that jointly performs the model compression (\ie, quantization) and the loss landscape smoothing}.

\item
We provide a unified view for the landscape smoothing of the quantized models, where we consider quantization and SAM as introducing quantization noises and adversarial perturbations to the model weights, respectively. Relying on this, we present three cases of SAQ
and make comprehensive comparisons among them. We further introduce an efficient training strategy to largely reduce the computational overhead brought by SAQ while keeping its performance gain.

\item
Experiments on both CNNs and Transformers \revise{across a variety of datasets} show that \methodshortname improves quantized models' generalization performance and performs favorably against SOTA \jing{uniform} quantization methods. For example, on ImageNet, our 4-bit ViT-B/16 surpasses AdamW by 1.2\% \jing{on the} Top-1 accuracy. Moreover, our 4-bit ResNet-50 exceeds the SOTA method by 0.9\% on the Top-1 accuracy. 
\end{itemize} 

\section{Related Work}
\label{sec:related_work}
\noindent\textbf{Network quantization.} Network quantization seeks to reduce the model size and computational cost by mapping weights, activations, and even gradients of a CNN or ViT to low-precision ones. Existing quantization methods can be roughly divided into two categories according to the quantization bitwidth, namely, fixed-point quantization~\cite{zhou2016dorefa,cai2017deep,hou2018loss,choi2018pact,zhuang2018towards,zhang2018lq,jung2019learning,Esser2020LEARNED,chen2021aqd,kim2021bert,liu2021post,han2021improving} and binary quantization~\cite{hubara2016binarized,rastegari2016xnor,liu2018bi,lin2017towards,liu2021adam,bai2021binarybert,qin2021bibert}. To reduce the quantization error, %
existing methods~\cite{choi2018pact,zhang2018lq,jung2019learning,Esser2020LEARNED,bhalgat2020lsq+,yamamoto2021learnable} explicitly parameterize the quantizer and train it jointly with network parameters. 
To reduce the optimization difficulty incurred by the non-differentiable discretization, extensive methods~\cite{ding2019regularizing,Yang_2019_CVPR,gong2019differentiable,lee2021network,kim2021distance} have been proposed to approximate the gradients.
To encourage more information to be maintained by the quantized weights, several weight regularization methods~\cite{han2021improving,liu2022nonuniform} have been proposed to alleviate the discrepancy between the full-precision and low-precision weights. 
To improve robustness against different bitwidths of quantization, \cite{Alizadeh2020Gradient} introduces $\ell_1$-norm of the loss gradients as a regularization term.
Compared with these methods, \methodshortname focuses on improving the generalization performance of the quantized models from a new perspective by smoothing the loss landscape. \revise{Compared with recent studies~\cite{nagel2022overcoming,defossez2021differentiable} that mitigates oscillations incurred by implicit stochasticity of the straight-through estimator (STE)~\cite{bengio2013estimating} to stabilize optimization, \methodshortname instead reduces the effect of adversarial perturbations and quantization noises by directly minimizing the perturbed quantization loss and vanilla quantization loss.}

\noindent\textbf{Loss geometry and generalization.}
Hochreiter~\etal~\cite{hochreiter1995simplifying} pioneered the proposition that flat local minima may generalize better in neural networks. 
Following that, several studies have been proposed to investigate the relation between the geometry of the loss landscape and the generalization performance of the models~\cite{keskar2017large,smith2017bayesian,dziugaite2017computing,ChaudhariCSLBBC17,jiang2019fantastic,moosavi2019robustness,liu2020loss}. Recently, Sharpness-Aware Minimization (SAM)~\cite{foret2021sharpnessaware} seeks to find parameters that lie in a region with uniformly low loss value and shows promising performance across various architectures and benchmark datasets. 
Concurrent works have also been proposed to introduce adversarial weight perturbations to improve the robustness against adversarial examples~\cite{wu2020adversarial} or generalization performance~\cite{zheng2021regularizing}. 
However, the computational overhead of these methods is roughly doubled compared with those using conventional optimizers (\eg, SGD and AdamW). To address this issue, ESAM~\cite{du2021efficient}, LookSAM~\cite{liu2022towards}, SAF~\cite{du2022sharpness}, AE-SAM~\cite{jiang2023an} have been proposed to accelerate the SAM optimization without performance drop. Apart from the efficiency issues, 
several methods including
ASAM~\cite{Kwon2021ASAMAS}, GSAM~\cite{zhuang2021surrogate} and Fisher SAM~\cite{kim2022fisher}
have been proposed to improve the performance of SAM. More recently, SAM has been applied to improve the performance of the pruned models~\cite{na2022train}.
While these methods target on full-precision models,
our proposed \methodshortname focuses on improving the generalization performance of the quantized models, which \jing{is a pioneering one in the sense that we} jointly perform model compression (\ie, quantization) and loss landscape smoothing. %

\section{Preliminary}
\label{sec:preliminary}
\subsection{Network Quantization}
\label{sec:quantization}
In this paper, we use uniform quantization \jing{which is  hardware-friendly~\cite{zhou2016dorefa}.}
Given an $L$-layer deep model, let $w^l$ and $x^l$ be the weight and input activation \wrt~the $l$-th layer. %
For simplicity, we omit the layer index $l$ afterward. Before performing quantization, we first normalize $w$ and $x$ %
into the scale of $[0,1]$ by applying clipping as
\begin{equation}
\setlength\abovedisplayskip{3pt}%
\setlength\belowdisplayskip{3pt}
\begin{split}
    \hat{w} 
    & = \begin{cases}
    \frac{1}{2} \left( \frac{w}{\alpha_w} + 1 \right), & \text{if}~~ -1 < \frac{w}{\alpha_w} < 1 \\
    0, & \text{if}~~\frac{w}{\alpha_w} \le -1 \\
    1, & \text{if}~~\frac{w}{\alpha_w} \ge 1 \\
    \end{cases}, 
    ~\\
\hat{x} & = 
\begin{cases}
    \frac{x}{\alpha_x}, \hphantom{+1+1+}  & \text{if}~~0 < \frac{x}{\alpha_x} < 1 \\
    0, & \text{if}~~\frac{x}{\alpha_x} \le 0 \\
    1, & \text{if}~~\frac{x}{\alpha_x} \ge 1 \\
\end{cases},
\end{split}
\end{equation}
where $\alpha_w$ and $\alpha_x$ are layer-wise trainable clipping levels that limit the range of weight and activation, respectively. 
We then quantize $\hat{z} \in \{\hat{w}, \hat{x} \}$ to the discrete one $\bar{z} \in \{ \bar{w}, \bar{x} \}$ by $\bar{z} = D(\hat{z}, s) = s \cdot \lfloor {\hat{z}} / {s} \rceil$, where $\lfloor \cdot \rceil$ is a rounding operator that returns the nearest integer of a given value and $s=1/(2^b-1)$ is the normalized step size for $b$-bit quantization. \jing{Lastly}, we obtain the quantized $w$ and $x$ by
\begin{equation}
\setlength\abovedisplayskip{3pt}%
\setlength\belowdisplayskip{3pt}
\begin{split}
    Q_w(w) = \alpha_w ( 2 \bar{w} - 1), ~
    Q_x(x) = \alpha_x \bar{x}.
\end{split}
\end{equation}
During training, the rounding operation $\lfloor \cdot \rceil$ is non-differentiable. To overcome this issue, following~\cite{zhou2016dorefa,hubara2016binarized}, we apply the straight-through estimation (STE)~\cite{bengio2013estimating} to approximate the gradient of the rounding operator by identity mapping for \jing{backpropagation}, namely, $\partial \bar{z}/ \partial \hat{z} \approx 1$.

\subsection{Sharpness-Aware Minimization}
\label{sec:sam}
Without loss of generality, let $\mS=\{(\bx_i, y_i)\}_{i=1}^n$ be the training data.
The goal of model training is to minimize the empirical risk $\mL(\bw)=\frac{1}{n}\sum_{i=1}^n \ell(\bw,\bx_i, y_i)$, where $\ell(\bw,\bx_i, y_i)$ is a loss function for the sample $(\bx_i, y_i)$ with model weights $\bw$. Instead of seeking a single \jing{place with a local minimal loss}, Sharpness-Aware Minimization~\cite{foret2021sharpnessaware} (SAM) %
seeks a region that has uniformly low training loss (both low loss and low curvature), which can be formulated as a min-max optimization problem as
\begin{equation}
\setlength\abovedisplayskip{3pt}%
\setlength\belowdisplayskip{3pt}
    \label{eq:sam_objective}
    \min _{\bw} \max _{\left\| \boldsymbol{\epsilon} \right\|_{2} \leq \rho} \mL(\bw+\boldsymbol{\epsilon}),
\end{equation}
where 
the inner optimization problem attempts to find perturbations $\boldsymbol{\epsilon}$ in an $\ell_2$ Euclidean ball with a pre-defined radius $\rho$ that maximizes the perturbed loss $\mL(\bw+\boldsymbol{\epsilon})$. To solve the inner problem, SAM approximates the optimal $\boldsymbol{\epsilon}$ to maximize $\mL(\bw + \boldsymbol{\epsilon})$ using a first-order Taylor expansion as 
\begin{equation}
\setlength\abovedisplayskip{3pt}%
\setlength\belowdisplayskip{3pt}
\begin{split}
    \label{eq:compute_epsilon}
    \hat{\boldsymbol{\epsilon}} &= {\arg \max }_{\left\| \boldsymbol{\epsilon}\right\|_{2} \leq \rho} \mL(\bw+\boldsymbol{\epsilon}) \\
    &\approx {\arg \max}_{\left\| \boldsymbol{\epsilon}\right\|_{2} \leq \rho} \mL (\bw) + \boldsymbol{\epsilon}^{\top} \nabla_{\bw} \mL(\bw) \\
    &\approx \rho \frac{ \nabla_{\bw} \mL\left(\bw\right) }{\left\| \nabla_{\bw} \mL\left(\bw\right)\right\|_{2}}.
\end{split}
\end{equation}
By substituting Eq.~(\ref{eq:compute_epsilon}) back into Eq.~(\ref{eq:sam_objective}), we then have the following optimization problem:
\begin{equation}
\setlength\abovedisplayskip{3pt}%
\setlength\belowdisplayskip{3pt}
    \label{eq:outer_problem}
    \min _{\bw} \mL(\bw + \hat{\boldsymbol{\epsilon}} ).
\end{equation}
\jing{Lastly}, SAM updates the model weights based on the gradient $\nabla_{\bw} \mL(\bw) |_{\bw + \hat{\boldsymbol{\epsilon}}}$.

\section{Sharpness-Aware Quantization}
\label{sec:method}
As shown in Figure~\ref{fig:loss_landscape}, the low-precision model shows a much sharper loss landscape compared with the full-precision one. \jing{Therefore}, small perturbations on the full-precision weights may incur large changes in the quantized weights, leading to severe loss oscillation. As a result, the gradients are unstable during training, which \jing{might mislead} weight update and the resulting quantized model \jing{might} converge to poor local minima.
To overcome this, one may directly apply SAM to train the quantized models, which, however, \textit{can suffer from perturbation mismatch or diminishment problems due to the clipping and discretization operations in quantization (See Section~\ref{sec:saq_different_cases})}, resulting in suboptimal performance.

\revise{In the following, we describe our proposed Sharpness-Aware Quantization (\methodshortname) to smooth the loss landscape and improve the generalization performance of the quantized models. We begin with a unified view on the loss landscape smoothing of quantized models in Section~\ref{sec:unified_view_saq}, and then provide an analysis of three different cases of SAQ in Section~\ref{sec:saq_different_cases}. Last, we introduce a fast optimization method for SAQ in Section~\ref{sec:fast_optimization}.}

\subsection{Unifying Quantization and SAM}
\label{sec:unified_view_saq}
We consider quantization and SAM as introducing quantization noises $\boldsymbol{\epsilon}_q$ and adversarial perturbations $\boldsymbol{\epsilon}_s$ to the model weights $\bw$, respectively, which provides a unified view for the loss landscape smoothing of the quantized models. In this way, 
the optimization problem is
\begin{equation}
\setlength\abovedisplayskip{3pt}%
\setlength\belowdisplayskip{3pt}
\begin{split}
    \label{eq:unified_equation}
    \min_{\bw, \boldsymbol{\alpha}_w, \boldsymbol{\alpha}_x} \mL (\bw + \boldsymbol{\epsilon}_q + \hat{\boldsymbol{\epsilon}}_s) 
    ~
    \text{where}~\hat{\boldsymbol{\epsilon}}_s={\arg \max}_{\left\| \boldsymbol{\epsilon}_s \right\|_{2} \leq \rho} \mL_p,
\end{split}
\end{equation}
where $\mL (\bw + \boldsymbol{\epsilon}_q + \hat{\boldsymbol{\epsilon}}_s)$ is a perturbed quantization loss and $\mL_p$ is a perturbed loss \revise{depending on the full-precision weights $\bw$ or the quantized weights $Q_w(\bw)$.}

\subsection{Case Analysis for SAQ}
\label{sec:saq_different_cases}
To solve the optimization problem in Eq.~(\ref{eq:unified_equation}), we need to obtain $\boldsymbol{\epsilon}_q$ as well as $\hat{\boldsymbol{\epsilon}}_s$. According to whether $\boldsymbol{\epsilon}_q$ and $\hat{\boldsymbol{\epsilon}}_s$ are dependent on each other, we can %
transform the loss function in
Eq.~(\ref{eq:unified_equation}) to different objectives, as shown in Table~\ref{table:objective}.
\revise{To simplify the notation, we define}
the quantization error function $\boldsymbol{\epsilon}_q(\bw)$ and the perturbation function $\hat{\boldsymbol{\epsilon}}_s(\bw)$ as
\begin{equation}
\setlength\abovedisplayskip{3pt}%
\setlength\belowdisplayskip{3pt}
    \label{eq:quantization_noise}
    \boldsymbol{\epsilon}_q(\bw) = Q_w(\bw) - \bw,~
    \hat{\boldsymbol{\epsilon}}_s(\bw) = \rho \frac{ \nabla_{\bw} \mL\left(\bw\right) }{\left\| \nabla_{\bw} \mL\left(\bw\right)\right\|_{2}}.
\end{equation}

\begin{table}[!h]
\centering
\vspace{-4mm}
\caption{Objectives for different cases of SAQ.
}
\vspace{-2mm}
\label{table:objective}
\renewcommand\arraystretch{1.1}
\scalebox{0.85}{
\begin{tabular}{cccccc}
\toprule
Name & Objective function \\
\midrule
Unified & $\mL (\bw + \boldsymbol{\epsilon}_q + \hat{\boldsymbol{\epsilon}}_s)$ \\
Case 1 & $\mL (\bw + \boldsymbol{\epsilon}_q(\bw) + \hat{\boldsymbol{\epsilon}}_s(\bw))$ \\
Case 2 & $\mL ({( \bw + \hat{\boldsymbol{\epsilon}}_s(\bw) )} + \boldsymbol{\epsilon}_q( { \bw + \hat{\boldsymbol{\epsilon}}_s(\bw) } ))$ \\
Case 3 & $\mL ({  ( \bw + \boldsymbol{\epsilon}_q(\bw) ) } + 
    \hat{\boldsymbol{\epsilon}}_s({ \bw + 
    \boldsymbol{\epsilon}_q(\bw)}))$ \\
\bottomrule
\end{tabular}
}
\vspace{-2mm}
\end{table}

\noindent\textbf{Case 1:} We calculate the quantization noises $\boldsymbol{\epsilon}_q$ and optimal perturbations $\hat{\boldsymbol{\epsilon}}_s$ independently. In this case, the perturbed loss is defined as $\mL_p = \mL(\bw + \boldsymbol{\epsilon}_s)$. By maximizing the perturbed loss with $\ell_2$-norm constraint, the optimal perturbations can be approximated by $\hat{\boldsymbol{\epsilon}}_s(\bw)$. In this way, the optimization problem can be transformed to
\begin{equation}
\setlength\abovedisplayskip{3pt}%
\setlength\belowdisplayskip{3pt}
\min_{\bw, \boldsymbol{\alpha}_w, \boldsymbol{\alpha}_x} \mL (\bw + 
\boldsymbol{\epsilon}_q(\bw) + \hat{\boldsymbol{\epsilon}}_s(\bw)).
\end{equation}
With Eq.~(\ref{eq:quantization_noise}), we have $\bw + \boldsymbol{\epsilon}_q(\bw)=Q_w(\bw)$. Then, the above problem can be rewritten as
\begin{equation}
\setlength\abovedisplayskip{3pt}%
\setlength\belowdisplayskip{3pt}
    \min_{\bw, \boldsymbol{\alpha}_w, \boldsymbol{\alpha}_x} \mL (Q_w(\bw) + \hat{\boldsymbol{\epsilon}}_s(\bw)).
\end{equation}
In this case, the optimal perturbations introduced to the quantized weights $Q_w(\bw)$ depend on the gradient of the full-precision weights $\bw$. Using the chain rule, the full-precision weights' gradient can be computed by
\begin{equation}
\setlength\abovedisplayskip{3pt}%
\setlength\belowdisplayskip{3pt}
\begin{split}
    \frac{\partial \mL_p(\bw)}{\partial w_i} &= \frac{\partial \mL_p(\bw)}{\partial Q_w(w_i)} \frac{\partial Q_w(w_i)}{\partial w_i} \\
    &= 
    \begin{cases}
         \frac{\partial \mL_p(\bw)}{ \partial Q_w(w_i)} & \text{if}~ -1 \le \frac{w_i}{\alpha_w^l} \le 1 \\
         0 & \text{otherwise}
    \end{cases},
\end{split}
\end{equation}
where $w_i$ is the $i$-th element of $\bw$ for layer $l$ and $\alpha_w^l$ is the corresponding clipping level. Due to the clipping operation, the difference between the full-precision weights' gradient $\partial \mL_p(\bw) / \partial w_i$ and the quantized weights' gradient $\partial \mL_p (\bw) / \partial Q_w(w_i)$ results in a perturbation mismatch problem, which makes the training process noisy and might degrade the quantization performance. 

Besides, Case 1 assumes that $\boldsymbol{\epsilon}_q$ and $\hat{\boldsymbol{\epsilon}}_s$ are computed independently, which ignores the dependency between them. To address this issue, we \jing{introduce} another two cases of SAQ in the following.

\noindent\textbf{Case 2:} We first combine model weights with the optimal perturbations $\hat{\boldsymbol{\epsilon}}_s$ and then introduce the quantization noises $\boldsymbol{\epsilon}_q$ to the perturbed model weights. In this way, the optimization problem is transformed to 
\begin{equation}
\setlength\abovedisplayskip{3pt}%
\setlength\belowdisplayskip{3pt}
    \label{eq:case_2}
    \min_{\bw, \boldsymbol{\alpha}_w, \boldsymbol{\alpha}_x} \mL ({ ( \bw + \hat{\boldsymbol{\epsilon}}_s(\bw) )} + \boldsymbol{\epsilon}_q( { \bw + \hat{\boldsymbol{\epsilon}}_s(\bw) } )).
\end{equation}
Same as Case 1, the perturbed loss is $\mL_p = \mL(\bw + \boldsymbol{\epsilon}_s)$ and the optimal perturbations can be obtained by $\hat{\boldsymbol{\epsilon}}_s(\bw)$. In this case, the quantization noises $\boldsymbol{\epsilon}_q(\bw + \hat{\boldsymbol{\epsilon}}_s(\bw))$ is represented as a function of the optimal perturbations $\hat{\boldsymbol{\epsilon}}_s(\bw)$.
Using Eq.~(\ref{eq:quantization_noise}), we reformulate the problem as
\begin{equation}
\setlength\abovedisplayskip{3pt}%
\setlength\belowdisplayskip{3pt}
    \label{eq:reformulate_case_2}
    \min_{\bw, \boldsymbol{\alpha}_w, \boldsymbol{\alpha}_x} \mL (Q_w(\bw + \hat{\boldsymbol{\epsilon}}_s(\bw))).
\end{equation}
Nevertheless, the introduced \jing{small} perturbations may not change the resulting quantized weights due to the discretization process \revise{and results in perturbation diminishment issue}, \ie, $Q_w(\bw + \hat{\boldsymbol{\epsilon}}_s(\bw))=Q_w(\bw)$. As a result, $\mL (Q_w(\bw + \hat{\boldsymbol{\epsilon}}_s(\bw)))$ might be reduced to $\mL (Q_w(\bw))$, which degenerates to the regular quantization. 

\noindent\textbf{Case 3:} 
We first combine model weights with the quantization noises $\boldsymbol{\epsilon}_q$ and then introduce the optimal perturbations $\hat{\boldsymbol{\epsilon}}_s$. In this way, the optimization problem becomes
\begin{equation}
\setlength\abovedisplayskip{3pt}%
\setlength\belowdisplayskip{3pt}
    \label{eq:case_3}
    \min_{\bw, \boldsymbol{\alpha}_w, \boldsymbol{\alpha}_x} \mL ({  ( \bw + \boldsymbol{\epsilon}_q(\bw) ) } + 
    \hat{\boldsymbol{\epsilon}}_s({ \bw + 
    \boldsymbol{\epsilon}_q(\bw)})).
\end{equation}
In this case, we define the perturbed loss as $\mL_p=\mL(\bw + \boldsymbol{\epsilon}_q(\bw) + \boldsymbol{\epsilon}_s)$ and obtain the optimal perturbations by $\hat{\boldsymbol{\epsilon}}_s(\bw + \boldsymbol{\epsilon}_q(\bw))$ which is expressed as a function of the quantization noises $\boldsymbol{\epsilon}_q(\bw)$.
With Eq.~(\ref{eq:quantization_noise}), the optimization problem can be rewritten as %
\begin{equation}
\setlength\abovedisplayskip{3pt}%
\setlength\belowdisplayskip{3pt}
    \label{eq:reformulate_case_3}
    \min_{\bw, \boldsymbol{\alpha}_w, \boldsymbol{\alpha}_x} \mL (Q_w(\bw) + \hat{\boldsymbol{\epsilon}}_s(Q_w(\bw))),
\end{equation}
where we introduce perturbations to the quantized weights $Q_w(\bw)$ rather than the full-precision weights $\bw$ as in Case 2.
In this way, the introduced perturbations will not be diminished by the quantization operation.
Moreover, compared with Case 1, Case 3 does not suffer from the perturbation mismatch issue since the optimal perturbations depend on the gradient of the quantized weights instead of the full-precision ones.
In summary, Case 3 is the best suited to smooth the loss landscape of the quantized models.

\subsection{Fast Optimization for SAQ}
\label{sec:fast_optimization}
\noindent\textbf{Final objective.}
In SAQ, we seek model parameters $\bu \in \{ \bw, \boldsymbol{\alpha}_w, \boldsymbol{\alpha}_x \}$ that is located in a neighborhood with uniformly low value of $\mL(Q_w(\bw))$ by minimizing a surrogate loss $\mL(\bw + \boldsymbol{\epsilon}_q + \hat{\boldsymbol{\epsilon}}_s)$. However, the sharp loss landscape of the quantized model leads to a large angle $\theta$ between the gradient of the perturbed quantization loss $\bg_s = \nabla_{\bu} \mL\left(\bw+\boldsymbol{\epsilon}_q+\hat{\boldsymbol{\epsilon}}_s\right)$ and the gradient of vanilla quantization loss $\bg = \nabla_{\bu} \mL(Q_w(\bw))$ (See Figure~\ref{fig:cosine_distance_two_loss}), which forms a gap between $\mL(Q_w(\bw))$ and $\mL(\bw + \boldsymbol{\epsilon}_q + \hat{\boldsymbol{\epsilon}}_s)$, particularly at extreme low-bit cases. Consequently, the optimization becomes challenging and leads to sub-optimal performance.

To overcome this challenge, we introduce an additional vanilla quantization loss $\mL(Q_w(\bw))$ into the objective and reformulate the optimization problem as
\begin{equation}
\setlength\abovedisplayskip{3pt}%
\setlength\belowdisplayskip{3pt}
\begin{split}
    \label{eq:unified_equation_2}
    &\min_{\bw, \boldsymbol{\alpha}_w, \boldsymbol{\alpha}_x} \mL (\bw + \boldsymbol{\epsilon}_q + \hat{\boldsymbol{\epsilon}}_s) + \mL(Q_w(\bw)) \\
    &\text{where}~\hat{\boldsymbol{\epsilon}}_s={\arg \max}_{\left\| \boldsymbol{\epsilon} \right\|_{2} \leq \rho} \mL_p,
\end{split}
\end{equation}
where we enforce the quantized models to find minima with both low $\mL (\bw + \boldsymbol{\epsilon}_q + \hat{\boldsymbol{\epsilon}}_s)$ and $\mL(Q_w(\bw))$.
As the gradient of $\mL( Q_w(\bw) )$ has been computed during the backpropagation when solving the inner problem as shown in Eq.~(\ref{eq:compute_epsilon}), we can reuse them when solving the outer problem.

\begin{figure}[!t]
    \begin{center}
        \centering
        \includegraphics[height=1.0in]{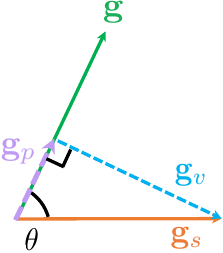}
        \vspace{-0.1in}
        \caption{An illustration of gradient decomposition in SAQ where we decompose ${\color{customorange}\bg_s = \nabla_{\bu} \mL\left(\bw+\boldsymbol{\epsilon}_q+\hat{\boldsymbol{\epsilon}}_s\right)}$ into components ${\color{custompurple}\bg_p}$ and ${\color{customblue}\bg_v}$ that are parallel and vertical to ${\color{customgreen}\bg = \nabla_{\bu} \mL(Q_w(\bw))}$. $\theta$ is the angle between ${\color{customorange}\bg_s}$ and ${\color{customgreen}\bg}$. }
        \label{fig:gradient_diagram}
    \end{center}
    \vspace{-0.4in}
\end{figure}

\noindent\textbf{Efficient training.}
Note that solving the problem in Eq.~(\ref{eq:unified_equation_2}) requires additional forward and backward propagation, which results in roughly doubled training cost compared with regular optimizers such as SGD. To address this issue, we customize an efficient training strategy following LookSAM~\cite{liu2022towards}. As shown in Figure~\ref{fig:gradient_diagram}, we decompose the perturbed quantization loss gradient $\bg_s = \nabla_{\bu} \mL\left(\bw+\boldsymbol{\epsilon}_q+\hat{\boldsymbol{\epsilon}}_s\right)$ into two components, $\bg_p$ and $\mathbf{g}_v$, which are parallel and vertical to the gradient of the vanilla quantization loss, $\bg = \nabla_{\bu} \mL(Q_w(\bw))$, respectively. We have a similar observation as in LookSAM that $\bg_v$ changes much slower than $\bg_p$ and $\bg_s$ during training. Therefore, we only calculate $\bg_s$ every $\tau$ iterations and obtain its vertical component $\mathbf{g}_v$. For the intermediate iterations, we reuse the direction of $\bg_v$ to approximate $\bg_s = \bg + \beta \| \bg \|  \frac{{\bg_v}}{\| {\bg_v} \|}$, where $\beta$ is a hyperparameter scaling the gradient. As a result, SAQ is only slightly slower than the regular optimizers (See Table~\ref{table:effect_efficient_SAQ}). We then update the model parameters using a gradient combination of $\bg_s + \bg$. The training process of SAQ is summarized in Algorithm~\ref{alg:looksaq}.

\begin{figure}[!t]
\vspace{-0.2in}
\begin{algorithm}[H]
\caption{Training algorithm for SAQ}
\label{alg:looksaq}
\begin{algorithmic}[1]
    \REQUIRE Training set $\mD$, model parameter $\bu \in \{ \bw, \boldsymbol{\alpha}_w, \boldsymbol{\alpha}_x \}$, learning rate $\eta$, update frequency $\tau$, total training iterations $T$, hyperparameter $\beta$.
    \FOR{$t \in \{ 1, 2, \dots, T\}$}
    \STATE Sample a batch of data $\mB_t$ from $\mD$
    \STATE Compute ${\color{customgreen}\bg = \nabla_{\bu} \mL(Q_w(\bw))}$ on $\mB_t$
    \IF{$t = \tau k, ~~k \in \mathbb{N}^{+}$}
        \STATE Compute $\boldsymbol{\epsilon}_q$ and $\hat{\boldsymbol{\epsilon}}_s$
        \STATE Compute ${\color{customorange}\bg_s = \nabla_{\bu} \mL\left(\bw+\boldsymbol{\epsilon}_q+\hat{\boldsymbol{\epsilon}}_s\right)}$ on $\mB_t$
        \STATE Project ${\color{customorange}\bg_s}$ onto ${\color{customgreen}\bg}$ to obtain ${\color{custompurple}\bg_p}= \frac{ {\color{customorange}\bg_s}^{\top} {\color{customgreen}\bg}}{ {\color{customgreen}\bg}^{\top} {\color{customgreen}\bg} } {\color{customgreen}\bg}$
        \STATE Compute ${\color{customblue}\bg_v} = {\color{customorange}\bg_s} - {\color{custompurple}\bg_p} $
    \ELSE
        \STATE Compute ${\color{customorange}\bg_s} = {\color{customgreen}\bg} + \beta \| {\color{customgreen}\bg} \|  \frac{{\color{customblue}\bg_v}}{\| {\color{customblue}\bg_v} \|} $
    \ENDIF
    \STATE Update $\bu$ by $\bu = \bu - \eta ({\bg_s + \bg})$
    \ENDFOR
\end{algorithmic}
\end{algorithm}
\vspace{-0.42in}
\end{figure}

\section{Experiments}
\label{sec:experiments}
\begin{table}[t]
\renewcommand{\arraystretch}{1.3}
\caption{Performance comparisons of different methods \jing{with} ResNet-18, ResNet-34 and ResNet-50 on ImageNet. We obtain DoReFa-Net results from~\cite{choi2018pact}. ``W/A'' refers to the bitwidth of weights and activations, respectively. ``FP'' represents the Top-1 accuracy of the full-precision models. ``-'' denotes that the results are not reported. \revise{We do not apply advanced techniques to boost the performance as mentioned in the compared methods in Section~\ref{sec:experiments}.} 
}
\vspace{-0.1in}
\centering
\scalebox{0.72}
{
\begin{tabular}{cccccccc}
\toprule
\multirow{2}{*}{Method} & Bitwidth  & 
\multicolumn{2}{c}{Accuracy (\%)} & Bitwidth  & 
\multicolumn{2}{c}{Accuracy (\%)} \\
& (W/A) & Top-1 & Top-5 & (W/A) & Top-1 & Top-5 \\
\midrule
\multicolumn{7}{c}{ResNet-18 (FP: 70.7)} \\
\cdashline{1-7}
DoReFa-Net$^{*}$~\cite{zhou2016dorefa} & 4/4 & 68.1 & 88.1 & 2/2 & 62.6 & 84.4 \\
PACT$^{*}$~\cite{choi2018pact} & 4/4 & 69.2 & 89.0 & 2/2 & 64.4 & 85.6 \\
LQ-Nets$^{*}$~\cite{zhang2018lq} & 4/4 & 69.3 & 88.8 & 2/2 & 64.9 & 85.9 \\
DSQ~\cite{gong2019differentiable} & 4/4 & 69.6 & - & 2/2 & 65.2 & - \\
BRECQ~\cite{li2021brecq} & 4/4 & 69.6 & - & 2/2 & - & - \\
FAQ~\cite{mckinstry2019discovering} & 4/4 & 69.8 & 89.1 & 2/2 & - & - \\
QIL$^{*}$~\cite{jung2019learning} & 4/4 & 70.1 & - & 2/2 & 65.7 & - \\
LLT$^{*}$~\cite{wang2022learnable} & 4/4 & 70.4 & 89.6 & 2/2 & 66.0 & 86.2 \\
Auxi~\cite{zhuang2020training} & 4/4 & - & - & 2/2 & 66.7 & 87.0 \\
DAQ$^*$~\cite{kim2021distance} & 4/4 & 70.5 & - & 2/2 & 66.9 & - \\
EWGS$^*$~\cite{lee2021network} & 4/4 & 70.6 & -  & 2/2 & 67.0 & - \\
BR~\cite{han2021improving} & 4/4 & 70.8 & 89.6  & 2/2 & 67.2 & 87.3 \\
APOT~\cite{Li2020Additive} & 4/4 & 70.7 & 89.6  & 2/2 & 67.3 & 87.5 \\
LSQ~\cite{Esser2020LEARNED} & 4/4 & 71.1 & \textbf{90.0} & 2/2 & 67.6 & \textbf{87.6} \\
SAQ (Ours) & 4/4 & \textbf{71.6} & \textbf{90.0} & 2/2 & \textbf{67.8} & \textbf{87.6} \\
\midrule
\multicolumn{7}{c}{ResNet-34 (FP: 74.1)} \\
\cdashline{1-7}
LQ-Nets$^{*}$~\cite{zhang2018lq} & 4/4 & - & -  & 2/2 & 69.8 & 89.1 \\
DSQ~\cite{gong2019differentiable} & 4/4 & 72.8 & -  & 2/2 & 70.0 & - \\
FAQ~\cite{mckinstry2019discovering} & 4/4 & 73.3 & 91.3  & 2/2 & - & - \\
QIL$^{*}$~\cite{jung2019learning} & 4/4 & 73.7 & -  & 2/2 & 70.6 & - \\
APOT~\cite{Li2020Additive} & 4/4 & 73.8 & 91.6  & 2/2 & 70.9 & 89.7 \\
DAQ$^*$~\cite{kim2021distance} & 4/4 & 73.7 & -  & 2/2 & 71.0 & - \\
Auxi~\cite{zhuang2020training} & 4/4 & - & - & 2/2 & 71.2 & 89.8 \\
EWGS$^*$~\cite{lee2021network} & 4/4 & 73.9 & -  & 2/2 & 71.4 & - \\
LSQ~\cite{Esser2020LEARNED} & 4/4 & 74.1 & 91.7 & 2/2 & 71.6 & 90.3 \\
SAQ (Ours) & 4/4 & \textbf{75.0}& \textbf{92.3}  & 2/2 & \textbf{71.8} & \textbf{90.4} \\
\midrule
\multicolumn{7}{c}{ResNet-50 (FP: 76.8)} \\
\cdashline{1-7}
DoReFa-Net$^{*}$~\cite{zhou2016dorefa} & 4/4 & 71.4 & 89.8  & 2/2 & 67.1 & 87.3 \\
LQ-Net$^{*}$~\cite{zhang2018lq}  & 4/4 & 75.1 & 92.4  & 2/2 & 71.5 & 90.3\\
FAQ~\cite{mckinstry2019discovering} & 4/4 & 76.3 & 93.0  & 2/2 & - & - \\
PACT$^{*}$~\cite{choi2018pact}  & 4/4 & 76.5 & 93.2  & 2/2 & 72.2 & 90.5 \\
APOT~\cite{Li2020Additive} & 4/4 & 76.6 & 93.1  & 2/2 & 73.4 & 91.4 \\
LSQ~\cite{Esser2020LEARNED} & 4/4 & 76.7 & 93.2  & 2/2 & 73.7 & 91.5 \\
Auxi~\cite{zhuang2020training} & 4/4 & - & -  & 2/2 & 73.8 & 91.4 \\
SAQ (Ours) & 4/4 & \textbf{77.6} & \textbf{93.6} & 2/2 & \textbf{74.5} & \textbf{91.9} \\
\bottomrule
\end{tabular}
}
\begin{tablenotes}
     \item \footnotesize $^{*}$ denotes that the first and last layers are not quantized.
\end{tablenotes}
\label{table:results_on_imagenet_resnet}
\vspace{-0.3in}
\end{table}

\noindent\textbf{Datasets and evaluation metrics.} We evaluate our method on  ImageNet~\cite{deng2009imagenet} which is a large-scale dataset containing 1.28 million training images and 50k validation samples with 1k classes.
We measure the performance of different methods using the Top-1 and Top-5 accuracy. 

\noindent\textbf{Implementation details.}
Our implementations are based on PyTorch~\cite{paszke2019pytorch}. We apply \methodshortname to CNNs and vision Transformers, including ResNet-18~\cite{he2016deep}, ResNet-34, ResNet-50,  MobileNetV2~\cite{sandler2018inverted} and ViT~\cite{dosovitskiy2020image}. 
We first train the full-precision models and use them to initialize the low-precision ones. Following LSQ~\cite{Esser2020LEARNED}, we quantize both weights and activations for all matrix multiplication layers, including convolutional layers, fully-connected layers, and self-attention layers. For the first and last layers, we quantize both weights and activations to 8-bit to preserve the performance. \revise{We do not apply quantization to the input images since they have been quantized to 8-bit during image preprocessing.} 

For CNNs, we use the \jing{uniform} quantization method mentioned in Section~\ref{sec:quantization}. Relying on SGD with the momentum term of 0.9, we apply \methodshortname with Case 3 to train the quantized models with a mini-batch size of 512 unless otherwise specified. Following APOT~\cite{Li2020Additive}, we use weight normalization before quantization. We initialize the clipping levels to 1. We fine-tune 90 epochs for ResNet-18, ResNet-34, and ResNet-50. We set weight decay to $1\times10^{-4}$ by default except for 2-bit ResNet-18, for which we set it to $2.5\times10^{-5}$ following~\cite{Esser2020LEARNED}.
For MobileNetV2, we fine-tune 140 epochs 
\revise{following~\cite{park2020profit}}. We set the weight decay to $4\times10^{-5}$.
The learning rate is initialized to 0.02 and decreased to 0 following the cosine annealing~\cite{loshchilov2016sgdr}. For ViTs, we use LSQ+~\cite{bhalgat2020lsq+} \jing{uniform} quantization following Q-ViT~\cite{li2022q}. We initialize the clipping levels by minimizing the quantization error following~\cite{li2021fixed}. \revise{Relying on AdamW~\cite{loshchilov2018decoupled}, we apply SAQ with Case 3 to train ViTs.} The learning rate is initialized to $2 \times 10^{-4}$ and decreased to 0 using the cosine annealing. We train the quantized model for 150 epochs with a mini-batch size of 1,024. 
We do not apply the automatic mixed-precision training following~\cite{li2022q}. For the hyperparameter $\rho$, we conduct grid search over \{0.02, 0.05, 0.1, 0.15, 0.2, $\dots$, 1.0\} to find appropriate values following the common practice in SAM~\cite{foret2021sharpnessaware} and GSAM~\cite{zhuang2021surrogate}. More details regarding $\rho$ and its sensitivity analysis can be found in the supplementary material.
Following LookSAM~\cite{liu2022towards}, we set hyperparameter $\beta$ and update frequency $\tau$ to 0.7 and 4, respectively. 
\revise{Due to limited space, we put more implementation details in the supplementary material.}

\noindent\textbf{Compared methods.} 
We compare with enormous \jing{fixed-point} quantization methods, including 
DoReFa-Net~\cite{zhou2016dorefa}, PACT~\cite{choi2018pact}, LQ-Nets~\cite{zhang2018lq}, DSQ~\cite{gong2019differentiable},  FAQ~\cite{mckinstry2019discovering}, QIL~\cite{jung2019learning}, Auxi~\cite{zhuang2020training}, PROFIT~\cite{park2020profit}, LSQ~\cite{Esser2020LEARNED}, APOT~\cite{Li2020Additive},  LSQ+~\cite{bhalgat2020lsq+}, LLSQ~\cite{Zhao2020Linear}, DAQ~\cite{kim2021distance}, BRECQ~\cite{li2021brecq}, EWGS~\cite{lee2021network}, BR~\cite{han2021improving}, OOQ~\cite{nagel2022overcoming}, and LLT~\cite{wang2022learnable}.

\begin{table}[!t]
\renewcommand{\arraystretch}{1.3}
\caption{Performance comparisons in terms of ViT-S/32, ViT-S/16, ViT-B/32, ViT-B/16, and MobileNetV2 on ImageNet. We obtain the results of PACT from~\cite{wang2019haq}. \revise{We do not apply iterative training with weight freezing and progressive quantization in PROFIT~\cite{park2020profit} to improve the performance of the quantized models.}
}
\vspace{-0.1in}
\centering
\scalebox{0.75}
{
\begin{tabular}{cccccccc}
\toprule
Network & Method & \tabincell{c}{Bitwidth \\ (W/A)}  & \tabincell{c}{Top-1 \\ Acc. (\%)} & \tabincell{c}{Top-5 \\ Acc. (\%)} \\
\midrule
\multirow{2}{*}{\tabincell{c}{ViT-S/32 \\ (FP: 68.5)}} & LSQ+~\cite{bhalgat2020lsq+} & 4/4 & 68.0 & 88.1 \\
& SAQ (Ours) & 4/4 & \textbf{68.6} & \textbf{88.4} \\
\midrule
\multirow{2}{*}{\tabincell{c}{ViT-S/16 \\ (FP: 75.9)}} & LSQ+~\cite{bhalgat2020lsq+} & 4/4 & 76.1 & 93.0 \\
& SAQ (Ours) & 4/4 & \textbf{76.9} & \textbf{93.5} \\
\midrule
\multirow{2}{*}{\tabincell{c}{ViT-B/32 \\ (FP: 70.7)}} & LSQ+~\cite{bhalgat2020lsq+} & 4/4 & 72.1 & 90.4 \\
& SAQ (Ours) & 4/4 & \textbf{72.7} & \textbf{90.7} \\
\midrule
\multirow{2}{*}{\tabincell{c}{ViT-B/16 \\ (FP: 77.2)}} & LSQ+~\cite{bhalgat2020lsq+} & 4/4 & 78.0 & 93.4 \\
& SAQ (Ours) & 4/4 & \textbf{79.2} & \textbf{94.2} \\
\midrule
\multirow{9}{*}{\tabincell{c}{MobileNetV2 \\ (FP: 71.9)}} & PACT~\cite{choi2018pact} & 4/4 & 61.4 & 83.7 \\
& DSQ$^{*}$~\cite{gong2019differentiable} & 4/4 & 64.8 & - \\
& BRECQ~\cite{li2021brecq} & 4/4 & 66.6 & - \\
& LLSQ$^{*}$~\cite{Zhao2020Linear} & 4/4 & 67.4 & 88.0 \\
& EWGS~\cite{lee2021network} & 4/4 & 70.3 & - \\
& BR~\cite{han2021improving} & 4/4 & 70.4 & 89.4 \\
& OOQ~\cite{nagel2022overcoming} & 4/4 & 70.6 & - \\
& PROFIT~\cite{park2020profit} & 4/4 & 71.6 & \textbf{90.4} \\
& SAQ (Ours) & 4/4 & \textbf{72.0} & \textbf{90.4} \\
\bottomrule
\end{tabular}
}
\begin{tablenotes}
     \item \footnotesize $^{*}$ denotes that the first and last layers are not quantized.
\end{tablenotes}
\vspace{-0.25in}
\label{table:results_imagenet_mobilenet}
\end{table}

\revise{Unless otherwise specified, we do not apply advanced techniques to boost performance such as pre-activation in LSQ, non-uniform quantization in APOT, weight regularization in BR and OOQ, knowledge distillation in Auxi and PROFIT, iterative training with weight freezing in OOQ and PROFIT, gradient scaling in EWGS, progressive quantization in PROFIT, batch normalization re-estimation in PROFIT and OOQ, asymmetric quantization in PROFIT and LSQ+. More advanced techniques used in other quantization methods are discussed in the supplementary.}

\subsection{Main Results}
We apply \methodshortname to quantize ResNet-18, ResNet-34, and ResNet-50 on ImageNet. From Table~\ref{table:results_on_imagenet_resnet}, \methodshortname outperforms existing SOTA \jing{uniform quantization} methods by a large margin. The improvement is more obvious with the increase of bitwidth. For example, for 2-bit ResNet-34, the Top-1 accuracy improvement of \methodshortname over LSQ is 0.2\% while for the 4-bit one is 0.9\%. We speculate that the loss landscape of the quantized models becomes sharper with the decrease of bitwidths due to the discretization operation in quantization as shown in the supplementary. As a result, smoothing the loss landscapes of the 2-bit quantized models is harder than the 4-bit counterparts. Moreover, for 2-bit quantization, deeper models show more obvious accuracy improvement over the SOTA methods. For instance, \methodshortname surpasses Auxi by 0.7\% on 2-bit ResNet-50 while only bringing 0.2\% Top-1 accuracy improvement over LSQ on 2-bit ResNet-18.
\begin{table}[!t]
\renewcommand{\arraystretch}{1.3}
\caption{Performance comparisons of different cases. We report the results of ResNet-50 on ImageNet. $\lambda_{max}$ denotes the largest eigenvalue of the Hessian of the converged quantized model. Lower $\lambda_{\mathrm{max}}$ indicates flatter loss landscapes.}
\vspace{-0.1in}
\centering
\scalebox{0.7}
{
\begin{tabular}{cccccccccc}
\toprule
\multirow{2}{*}{Method} & Bitwidth  & 
\multicolumn{2}{c}{Acc. (\%)} & \multirow{2}{*}{$\lambda_{\mathrm{max}}$} & Bitwidth  & \multicolumn{2}{c}{Acc. (\%)} & \multirow{2}{*}{$\lambda_{\mathrm{max}}$} \\
& (W/A) & Top-1 & Top-5 & & (W/A) & Top-1 & Top-5 \\
\midrule
\multicolumn{9}{c}{\tabincell{c}{ResNet-50}} \\
\cdashline{1-9}
SGD & 4/4 & 76.5 & 93.1 & 71.8 & 2/2 & 73.9 & 91.6 & 60.1 \\
Case 1 & 4/4 & 77.3 & 93.5 & 6.6  & 2/2 & 74.3 & 91.8 & 12.6 \\
Case 2 & 4/4 & 77.0 & 93.3 & 14.0 & 2/2 & 74.2 & 91.8 & 24.4 \\
Case 3 & 4/4 & \textbf{77.6} & \textbf{93.6} & \textbf{6.3} & 2/2 & \textbf{74.5} & \textbf{91.9} & \textbf{9.5} \\
\bottomrule
\end{tabular}
}
\vspace{-0.12in}
\label{table:comparisons_different_cases}
\end{table}
Remarkably, our 4-bit ResNet-34 surpasses the full-precision model by 0.9\% on the Top-1 accuracy. One possible reason is that performing quantization with \methodshortname helps to remove redundancy and regularize the networks. Similar phenomena are also observed in LSQ. 

To show the effectiveness of our method on lightweight models, we apply \methodshortname to quantize MobileNetV2. From Table~\ref{table:results_imagenet_mobilenet}, our \methodshortname yields better performance than the SOTA \jing{uniform quantization} methods. For example, SAQ exceeds PROFIT by 0.4\% on the Top-1 accuracy. We also apply \methodshortname to ViT~\cite{dosovitskiy2020image}. We implement LSQ+ following~\cite{li2022q} and compare our method with it. From Table~\ref{table:results_imagenet_mobilenet}, our \methodshortname shows consistently superior performance over the baseline LSQ+ (\eg, 1.2\% Top-1 accuracy improvement on ViT-B/16).

\begin{figure}[!t]

        \begin{minipage}[t]{0.22\textwidth}
        \centering
        \includegraphics[height=1.0in]{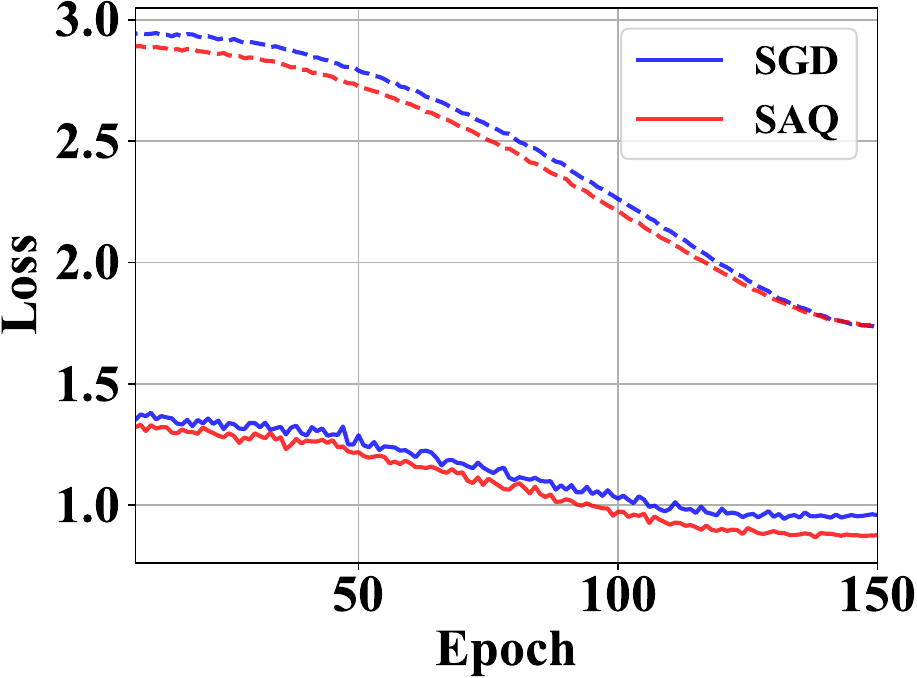}
        \vspace{-0.1in}
        \caption{\revise{The training (dashed line) and validation (solid line) losses for 4-bit ViT-B/16 on ImageNet. The $\lambda_{max}$ of SGD and SAQ obtained quantized models are 14,564 and 676, respectively.}}
        \label{fig:curve_comparisons}
        \end{minipage}
        \hspace{0.05in}
        \begin{minipage}[t]{0.23\textwidth}
        \centering
        \includegraphics[height=1.0in]{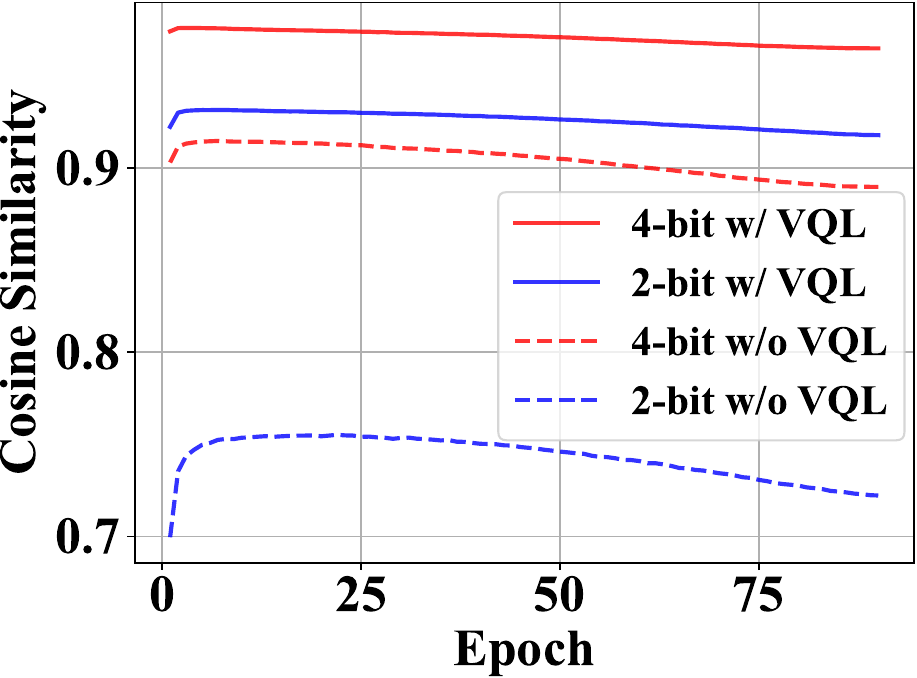}
        \vspace{-0.1in}
        \caption{Effect of introducing the vanilla quantization loss (VQL). We visualize the cosine similarity between $\nabla_{\bu} \mL(\bw + \boldsymbol{\epsilon}_q + \hat{\boldsymbol{\epsilon}}_s)$ and $\nabla_{\bu} \mL( Q_w(\bw) )$ in terms of 2-bit and 4-bit ResNet-18 on ImageNet.}
        \label{fig:cosine_distance_two_loss}
        \end{minipage}
    \vspace{-0.3in}
\end{figure}

\subsection{Ablation Studies}
\label{sec:ablation_studies}
\noindent\textbf{Performance comparisons of different cases.} To investigate the effectiveness of different cases introduced in Section~\ref{sec:saq_different_cases}, we apply different methods to quantize ResNet-50 on ImageNet. We use ``SGD'' to represent training the quantized models with the vanilla SGD. 
To measure the loss curvature, we report the largest eigenvalue $\lambda_{\mathrm{max}}$ of the Hessian of the converged quantized models following~\cite{chen2022when,foret2021sharpnessaware}. From Table~\ref{table:comparisons_different_cases}, Case 1, Case 2, and Case 3 all yield significantly higher accuracy and lower $\lambda_{\mathrm{max}}$ than the SGD counterpart. This strongly shows that our method is able to smooth the loss landscape and improve the generalization performance of the quantized models. Among the three cases, Case 2 performs the worst with the lowest accuracy and \jing{the} highest $\lambda_{\mathrm{max}}$, which \jing{suggests} that the perturbations introduced by SAM might be diminished due to the discretization, leading to suboptimal performance. Moreover, Case 3 consistently performs better than Case 1. For example, on 4-bit ResNet-50, Case 3 exceeds Case 1 by 0.3\% on the Top-1 accuracy \jing{as well as achieving lower $\lambda_{\mathrm{max}}$}. These results \jing{indicate} that the perturbation mismatch issue in Case 1 \jing{might degrade} the quantization performance.

\begin{table}[!t]
\renewcommand{\arraystretch}{1.3}
\caption{Effect of different losses in the objective function in Eq.~(\ref{eq:unified_equation_2}) on ImageNet.  ``VQL'' represents the vanilla quantization loss $\mL(Q_w(\bw))$ and ``PQL'' denotes the perturbed quantization loss $\mL (\bw + \boldsymbol{\epsilon}_q + \hat{\boldsymbol{\epsilon}}_s)$. 
}
\vspace{-0.1in}
\centering
\scalebox{0.72}
{
\begin{tabular}{cccccccccc}
\toprule
\multirow{2}{*}{VQL} & \multirow{2}{*}{PQL} & Bitwidth  & \multicolumn{2}{c}{Accuracy (\%)} & Bitwidth  & \multicolumn{2}{c}{Accuracy (\%)} \\
& & (W/A) & Top-1 & Top-5 & (W/A) & Top-1 & Top-5 \\
\midrule
\multicolumn{8}{c}{ResNet-18} \\
\cdashline{1-8}
\checkmark & & 2/2 & 67.3 & 87.4 & 4/4 & 71.1 & 89.8 \\
& \checkmark & 2/2 & 67.5 & 87.5 & 4/4 & 71.3 & 90.0 \\
\checkmark & \checkmark & 2/2 & \textbf{67.8} & \textbf{87.6} & 4/4 & \textbf{71.6} & \textbf{90.1}\\
\bottomrule
\end{tabular}
}
\label{table:effect_different_components}
\vspace{-0.25in}
\end{table}

\revise{Besides, to investigate the generalization capability of SGD and our SAQ, we show the training and validation losses of 4-bit ViT-B/16 on ImageNet in Figure~\ref{fig:curve_comparisons}. Compared with SGD, SAQ achieves lower training and validation losses at the beginning of training. At the end of training, SAQ shows slightly higher training loss but much lower validation loss and $\lambda_{{\mathrm{max}}}$ than SGD. These results justify that \methodshortname converges to a flatter local minimum and thus achieves better generalization performance. 
}

\noindent\textbf{Effect of different losses in the objective function.} To investigate the effect of different components in the objective in Eq.~(\ref{eq:unified_equation_2}), we apply different methods to quantize ResNet-18. 
From Table~\ref{table:effect_different_components}, using the perturbed quantization loss $\mL (\bw + \boldsymbol{\epsilon}_q + \hat{\boldsymbol{\epsilon}}_s)$ surpasses the one equipped with the vanilla quantization loss $\mL(Q_w(\bw))$ by 0.2\% on the Top-1 accuracy for 2-bit and 4-bit quantization, supporting that smoothing the loss landscape improves the generalization performance of the quantized models. By combining $\mL(Q_w(\bw))$ and $\mL (\bw + \boldsymbol{\epsilon}_q + \hat{\boldsymbol{\epsilon}}_s)$, we observe Top-1 accuracy improvement of 0.3\% for both 2-bit and 4-bit quantization. 

To further show the effect of $\mL(Q_w(\bw))$, we visualize the cosine similarity between $\nabla_{\bu} \mL(\bw + \boldsymbol{\epsilon}_q + \hat{\boldsymbol{\epsilon}}_s)$ and $\nabla_{\bu} \mL( Q_w(\bw) )$ of the quantized ResNet-18 in Figure~\ref{fig:cosine_distance_two_loss}. We observe that the cosine similarities of 2-bit models are lower than those of 4-bit ones during training. By introducing $\mL( Q_w(\bw) )$, all quantized models achieve higher similarities. These results justify that introducing $\mL(Q_w(\bw))$ helps to reduce the gap between two losses and improve the performance of the quantized models.

\begin{table}[t]
\renewcommand{\arraystretch}{1.3}
\caption{Effect of the efficient training (ET) strategy on ImageNet. We measure the training throughput on 4 NVIDIA V100 GPUs with a mini-batch size of 512.
}
\vspace{-0.1in}
\centering
\scalebox{0.72}
{
\begin{tabular}{ccccc}
\toprule
\multirow{2}{*}{Network} & \multirow{2}{*}{Method} & \multicolumn{2}{c}{Accuracy (\%)} & Train Throughput \\
& & Top-1 & Top-5 & (images/s) \\
\midrule
\multirow{3}{*}{4-bit ResNet-34} & SGD & 74.4 & 91.9 & \textbf{1632} \\
& SAQ w/o ET & \textbf{75.1} & 92.2 & 841 \\
& SAQ w/ ET & 75.0 & \textbf{92.3} & 1454 \\
\bottomrule
\end{tabular}
}
\label{table:effect_efficient_SAQ}
\vspace{-0.12in}
\end{table}

\noindent\textbf{Effect of the efficient training strategy.} To investigate the effect of the efficient training strategy mentioned in Section~\ref{sec:fast_optimization}, we apply SAQ to train 4-bit ResNet-34 with and without the efficient training strategy on ImageNet. The training throughput is quantified by the number of processed images per second on 4 NVIDIA V100 GPUs with a mini-batch size of 512. From Table~\ref{table:effect_efficient_SAQ}, we observe that our SAQ with efficient training strategy is only $\sim$11\% slower than SGD while achieving nearly the same performance as SAQ without the efficient training strategy.

\begin{table}[!t]
\renewcommand{\arraystretch}{1.3}
\caption{Effect of jointly performing quantization and loss landscape smoothing on ImageNet.
The Top-1 accuracy of the full-precision ResNet-18 obtained by SAM is 70.9\%. 
}
\vspace{-0.1in}
\centering
\scalebox{0.72}
{
\begin{tabular}{cccccccc}
\toprule
\multirow{2}{*}{Method} & Bitwidth  & 
\multicolumn{2}{c}{Accuracy (\%)} & Bitwidth  & 
\multicolumn{2}{c}{Accuracy (\%)} \\
& (W/A) & Top-1 & Top-5 & (W/A) & Top-1 & Top-5 \\
\midrule
\multicolumn{7}{c}{ResNet-18 (FP: 70.7)} \\
\cdashline{1-7}
SGD & 2/2 & 67.3 & 87.4 & 4/4 & 71.1 & 89.8 \\
SAM $\rightarrow$ SGD & 2/2 & 67.4 & 87.5 & 4/4 & 71.1 & 89.9 \\
SAQ (Ours) & 2/2 & \textbf{67.8} & \textbf{87.6} & 4/4 & \textbf{71.6} & \textbf{90.1} \\
\bottomrule
\end{tabular}
}
\label{table:SAQ_vs_flat_quantize}
\vspace{-0.25in}
\end{table}

\noindent\textbf{SAQ vs. train flat and then quantize.}
\revise{
To further investigate the effectiveness of \methodshortname, we compare our method with ``SAM $\rightarrow$ SGD'' that first obtains a full-precision model with SAM and then trains a quantized model with SGD using full-precision model weights as initialization. 
We also include ``SGD'' that trains the quantized models with SGD for comparisons. We report the results of different methods with ResNet-18 on ImageNet.
As seen from Table~\ref{table:SAQ_vs_flat_quantize}, for the full-precision model, SAM outperforms SGD by 0.2\% on the Top-1 accuracy. However, for 2-bit quantization, SAM $\rightarrow$ SGD only yields 0.1\% Top-1 accuracy improvement compared with SGD.}
We speculate that smoothing the loss landscape of the pre-trained models provides a better weight initialization for the quantized models.
Nevertheless, due to the large distribution gap between the quantized weights and full-precision weights, the performance gain over SGD is limited.
In contrast, SAQ performs %
better than SAM $\rightarrow$ SGD, which shows the superiority of jointly performing quantization and loss landscape smoothing. For example, on ResNet-18, SAQ exceeds SAM $\rightarrow$ SGD by 0.4\% on the Top-1 accuracy.
These results suggest that the improvement comes not only from SAM but also from our SAM customization for network quantization.

\begin{table}[t]
\renewcommand{\arraystretch}{1.3}
\caption{Transfer performance comparisons on downstream tasks. \revise{We measure the performance of different methods on 4-bit ResNet-50 using the Top-1 accuracy (\%).}
}
\vspace{-0.1in}
\centering
\scalebox{0.72}
{
\begin{tabular}{ccc}
\toprule
Method & SGD & SAQ (Ours) \\
\midrule
CIFAR-10~\cite{krizhevsky2009learning} & 97.0$\pm$0.0 & \textbf{97.1$\pm$0.1} \\
CIFAR-100~\cite{krizhevsky2009learning} & 82.4$\pm$0.2 & \textbf{83.1$\pm$0.2} \\
Oxford Flowers-102~\cite{parkhi2012cats} & 96.1$\pm$0.2 & \textbf{96.4$\pm$0.4} \\
Oxford-IIIT Pets~\cite{nilsback2008automated} & 94.9$\pm$0.2 & \textbf{95.9$\pm$0.2} \\
\bottomrule
\end{tabular}
}
\label{table:transfer_results}
\vspace{-0.22in}
\end{table}

\noindent\textbf{More results on transfer learning.}
To evaluate the transfer power of different quantized models, we conduct transfer learning \jing{experiments on new} datasets, including CIFAR-10~\cite{krizhevsky2009learning}, CIFAR-100, Oxford-IIIT Pets~\cite{parkhi2012cats}, and Oxford Flowers-102~\cite{nilsback2008automated}. We use the quantized models trained on ImageNet to initialize the model weights and then fine-tune all layers using SGD. %
We repeat the experiments 5 times and report the mean as well as the standard deviation of the Top-1 accuracy. More implementation details can be found in the supplementary. From Table~\ref{table:transfer_results}, \methodshortname leads to much better transfer performance. For example, on Oxford-IIIT Pets, SAQ quantized 4-bit ResNet-50 brings 1.0\% Top-1 accuracy improvement over the SGD counterpart. These results justify that \methodshortname improves the generalization performance by smoothing the loss landscape of the quantized models.

\section{Conclusion and Future Work}
\label{sec:conclusion}
In this paper, we have devised a new training approach, called Sharpness-Aware Quantization (\methodshortname), to improve the generalization capability of the quantized models, which jointly performs compression (\ie, quantization) and loss landscape smoothing for the first time. To this end, we have provided a unified view for the loss landscape smoothing of the quantized models by formulating quantization and SAM as introducing quantization noises and adversarial perturbations to the model weights. According to whether the quantization noises and adversarial perturbations are dependent on each other, we have formulated SAQ into three cases, which have been fully studied and compared. We have further introduced an efficient training strategy that substantially reduces the training overhead, allowing SAQ to achieve comparable training speed to that of the default optimization method. Extensive experiments on various datasets with different architectures including %
CNNs and Transformers have demonstrated that SAQ consistently improves the performance of the quantized models and yields the SOTA uniform quantization results. In the future, our method can be extended to jointly perform pruning, quantization, and loss landscape smoothing to obtain more compact models with better performance.

{\small
\bibliographystyle{ieee_fullname}
\bibliography{egbib}

\begin{thebibliography}{10}\itemsep=-1pt

\bibitem{Alizadeh2020Gradient}
Milad Alizadeh, Arash Behboodi, Mart van Baalen, Christos Louizos, Tijmen
  Blankevoort, and Max Welling.
\newblock Gradient $\ell_1$ regularization for quantization robustness.
\newblock In {\em ICLR}, 2020.

\bibitem{bai2021binarybert}
Haoli Bai, Wei Zhang, Lu Hou, Lifeng Shang, Jin Jin, Xin Jiang, Qun Liu,
  Michael~R Lyu, and Irwin King.
\newblock Binarybert: Pushing the limit of bert quantization.
\newblock In {\em ACL/IJCNLP}, 2021.

\bibitem{bengio2013estimating}
Yoshua Bengio, Nicholas L{\'e}onard, and Aaron Courville.
\newblock Estimating or propagating gradients through stochastic neurons for
  conditional computation.
\newblock {\em arXiv preprint arXiv:1308.3432}, 2013.

\bibitem{bhalgat2020lsq+}
Yash Bhalgat, Jinwon Lee, Markus Nagel, Tijmen Blankevoort, and Nojun Kwak.
\newblock Lsq+: Improving low-bit quantization through learnable offsets and
  better initialization.
\newblock In {\em CVPRW}, pages 696--697, 2020.

\bibitem{cai2017deep}
Zhaowei Cai, Xiaodong He, Jian Sun, and Nuno Vasconcelos.
\newblock Deep learning with low precision by half-wave gaussian quantization.
\newblock In {\em CVPR}, pages 5918--5926, 2017.

\bibitem{carion2020end}
Nicolas Carion, Francisco Massa, Gabriel Synnaeve, Nicolas Usunier, Alexander
  Kirillov, and Sergey Zagoruyko.
\newblock End-to-end object detection with transformers.
\newblock In {\em ECCV}, pages 213--229. Springer, 2020.

\bibitem{ChaudhariCSLBBC17}
Pratik Chaudhari, Anna Choromanska, Stefano Soatto, Yann LeCun, Carlo Baldassi,
  Christian Borgs, Jennifer~T. Chayes, Levent Sagun, and Riccardo Zecchina.
\newblock Entropy-sgd: Biasing gradient descent into wide valleys.
\newblock In {\em ICLR}, 2017.

\bibitem{chen2021aqd}
Peng Chen, Jing Liu, Bohan Zhuang, Mingkui Tan, and Chunhua Shen.
\newblock Aqd: Towards accurate quantized object detection.
\newblock In {\em CVPR}, pages 104--113, 2021.

\bibitem{chen2022when}
Xiangning Chen, Cho-Jui Hsieh, and Boqing Gong.
\newblock When vision transformers outperform resnets without pre-training or
  strong data augmentations.
\newblock In {\em ICLR}, 2022.

\bibitem{choi2018pact}
Jungwook Choi, Zhuo Wang, Swagath Venkataramani, Pierce I-Jen Chuang,
  Vijayalakshmi Srinivasan, and Kailash Gopalakrishnan.
\newblock Pact: Parameterized clipping activation for quantized neural
  networks.
\newblock {\em arXiv preprint arXiv:1805.06085}, 2018.

\bibitem{defossez2021differentiable}
Alexandre D{\'e}fossez, Yossi Adi, and Gabriel Synnaeve.
\newblock Differentiable model compression via pseudo quantization noise.
\newblock {\em TMLR}, 2022.

\bibitem{deng2009imagenet}
Jia Deng, Wei Dong, Richard Socher, Li-Jia Li, Kai Li, and Li Fei-Fei.
\newblock Imagenet: A large-scale hierarchical image database.
\newblock In {\em CVPR}, pages 248--255, 2009.

\bibitem{devlin2018bert}
Jacob Devlin, Ming{-}Wei Chang, Kenton Lee, and Kristina Toutanova.
\newblock {BERT:} pre-training of deep bidirectional transformers for language
  understanding.
\newblock In {\em NAACL-HLT}, pages 4171--4186, 2019.

\bibitem{ding2019regularizing}
Ruizhou Ding, Ting-Wu Chin, Zeye Liu, and Diana Marculescu.
\newblock Regularizing activation distribution for training binarized deep
  networks.
\newblock In {\em CVPR}, pages 11408--11417, 2019.

\bibitem{dong2019hawq}
Zhen Dong, Zhewei Yao, Amir Gholami, Michael~W Mahoney, and Kurt Keutzer.
\newblock Hawq: Hessian aware quantization of neural networks with
  mixed-precision.
\newblock In {\em ICCV}, pages 293--302, 2019.

\bibitem{dosovitskiy2020image}
Alexey Dosovitskiy, Lucas Beyer, Alexander Kolesnikov, Dirk Weissenborn,
  Xiaohua Zhai, Thomas Unterthiner, Mostafa Dehghani, Matthias Minderer, Georg
  Heigold, Sylvain Gelly, et~al.
\newblock An image is worth 16x16 words: Transformers for image recognition at
  scale.
\newblock In {\em ICLR}, 2021.

\bibitem{du2021efficient}
Jiawei Du, Hanshu Yan, Jiashi Feng, Joey~Tianyi Zhou, Liangli Zhen, Rick
  Siow~Mong Goh, and Vincent Tan.
\newblock Efficient sharpness-aware minimization for improved training of
  neural networks.
\newblock In {\em ICLR}, 2022.

\bibitem{du2022sharpness}
Jiawei Du, Daquan Zhou, Jiashi Feng, Vincent~YF Tan, and Joey~Tianyi Zhou.
\newblock Sharpness-aware training for free.
\newblock In {\em NeurIPS}, 2022.

\bibitem{dziugaite2017computing}
Gintare~Karolina Dziugaite and Daniel~M. Roy.
\newblock Computing nonvacuous generalization bounds for deep (stochastic)
  neural networks with many more parameters than training data.
\newblock In {\em UAI}, 2017.

\bibitem{Esser2020LEARNED}
Steven~K. Esser, Jeffrey~L. McKinstry, Deepika Bablani, Rathinakumar Appuswamy,
  and Dharmendra~S. Modha.
\newblock Learned step size quantization.
\newblock In {\em ICLR}, 2020.

\bibitem{foret2021sharpnessaware}
Pierre Foret, Ariel Kleiner, Hossein Mobahi, and Behnam Neyshabur.
\newblock Sharpness-aware minimization for efficiently improving
  generalization.
\newblock In {\em ICLR}, 2021.

\bibitem{gong2019differentiable}
Ruihao Gong, Xianglong Liu, Shenghu Jiang, Tianxiang Li, Peng Hu, Jiazhen Lin,
  Fengwei Yu, and Junjie Yan.
\newblock Differentiable soft quantization: Bridging full-precision and low-bit
  neural networks.
\newblock In {\em ICCV}, pages 4852--4861, 2019.

\bibitem{han2021improving}
Tiantian Han, Dong Li, Ji Liu, Lu Tian, and Yi Shan.
\newblock Improving low-precision network quantization via bin regularization.
\newblock In {\em ICCV}, pages 5261--5270, 2021.

\bibitem{he2016deep}
Kaiming He, Xiangyu Zhang, Shaoqing Ren, and Jian Sun.
\newblock Deep residual learning for image recognition.
\newblock In {\em CVPR}, pages 770--778, 2016.

\bibitem{hochreiter1995simplifying}
Sepp Hochreiter and J{\"u}rgen Schmidhuber.
\newblock Simplifying neural nets by discovering flat minima.
\newblock In {\em NeurIPS}, pages 529--536, 1995.

\bibitem{hou2018loss}
Lu Hou and James~T. Kwok.
\newblock Loss-aware weight quantization of deep networks.
\newblock In {\em ICLR}, 2018.

\bibitem{hubara2016binarized}
Itay Hubara, Matthieu Courbariaux, Daniel Soudry, Ran El-Yaniv, and Yoshua
  Bengio.
\newblock Binarized neural networks.
\newblock {\em NeurIPS}, 29, 2016.

\bibitem{jiang2023an}
Weisen Jiang, Hansi Yang, Yu Zhang, and James Kwok.
\newblock An adaptive policy to employ sharpness-aware minimization.
\newblock In {\em ICLR}, 2023.

\bibitem{jiang2019fantastic}
Yiding Jiang, Behnam Neyshabur, Hossein Mobahi, Dilip Krishnan, and Samy
  Bengio.
\newblock Fantastic generalization measures and where to find them.
\newblock In {\em ICLR}, 2020.

\bibitem{jung2019learning}
Sangil Jung, Changyong Son, Seohyung Lee, Jinwoo Son, Jae-Joon Han, Youngjun
  Kwak, Sung~Ju Hwang, and Changkyu Choi.
\newblock Learning to quantize deep networks by optimizing quantization
  intervals with task loss.
\newblock In {\em CVPR}, pages 4350--4359, 2019.

\bibitem{keskar2017large}
Nitish~Shirish Keskar, Dheevatsa Mudigere, Jorge Nocedal, Mikhail Smelyanskiy,
  and Ping Tak~Peter Tang.
\newblock On large-batch training for deep learning: Generalization gap and
  sharp minima.
\newblock In {\em ICLR}, 2017.

\bibitem{kim2021distance}
Dohyung Kim, Junghyup Lee, and Bumsub Ham.
\newblock Distance-aware quantization.
\newblock In {\em ICCV}, pages 5271--5280, 2021.

\bibitem{kim2022fisher}
Minyoung Kim, Da Li, Shell~X Hu, and Timothy Hospedales.
\newblock Fisher sam: Information geometry and sharpness aware minimisation.
\newblock In {\em ICML}, pages 11148--11161, 2022.

\bibitem{kim2021bert}
Sehoon Kim, Amir Gholami, Zhewei Yao, Michael~W Mahoney, and Kurt Keutzer.
\newblock I-bert: Integer-only bert quantization.
\newblock In {\em ICML}, pages 5506--5518. PMLR, 2021.

\bibitem{krizhevsky2009learning}
Alex Krizhevsky and Geoffrey Hinton.
\newblock Learning multiple layers of features from tiny images.
\newblock {\em Tech Report}, 2009.

\bibitem{Kwon2021ASAMAS}
Jungmin Kwon, Jeongseop Kim, Hyun-Seok Park, and In~Kwon Choi.
\newblock Asam: Adaptive sharpness-aware minimization for scale-invariant
  learning of deep neural networks.
\newblock In {\em ICML}, 2021.

\bibitem{lee2021network}
Junghyup Lee, Dohyung Kim, and Bumsub Ham.
\newblock Network quantization with element-wise gradient scaling.
\newblock In {\em CVPR}, pages 6448--6457, 2021.

\bibitem{Li2018VisualizingTL}
Hao Li, Zheng Xu, Gavin Taylor, and Tom Goldstein.
\newblock Visualizing the loss landscape of neural nets.
\newblock In {\em NeurIPS}, 2018.

\bibitem{Li2020Additive}
Yuhang Li, Xin Dong, and Wei Wang.
\newblock Additive powers-of-two quantization: An efficient non-uniform
  discretization for neural networks.
\newblock In {\em ICLR}, 2020.

\bibitem{li2021brecq}
Yuhang Li, Ruihao Gong, Xu Tan, Yang Yang, Peng Hu, Qi Zhang, Fengwei Yu, Wei
  Wang, and Shi Gu.
\newblock Brecq: Pushing the limit of post-training quantization by block
  reconstruction.
\newblock In {\em ICLR}, 2021.

\bibitem{li2021fixed}
Zhexin Li, Peisong Wang, Zhiyuan Wang, and Jian Cheng.
\newblock Fixed-point quantization for vision transformer.
\newblock In {\em CAC}, pages 7282--7287. IEEE, 2021.

\bibitem{li2022q}
Zhexin Li, Tong Yang, Peisong Wang, and Jian Cheng.
\newblock Q-vit: Fully differentiable quantization for vision transformer.
\newblock {\em arXiv preprint arXiv:2201.07703}, 2022.

\bibitem{lin2017towards}
Xiaofan Lin, Cong Zhao, and Wei Pan.
\newblock Towards accurate binary convolutional neural network.
\newblock In {\em NeurIPS}, pages 345--353, 2017.

\bibitem{liu2020loss}
Chen Liu, Mathieu Salzmann, Tao Lin, Ryota Tomioka, and Sabine S{\"{u}}sstrunk.
\newblock On the loss landscape of adversarial training: Identifying challenges
  and how to overcome them.
\newblock In {\em NeurIPS}, 2020.

\bibitem{liu2022towards}
Yong Liu, Siqi Mai, Xiangning Chen, Cho-Jui Hsieh, and Yang You.
\newblock Towards efficient and scalable sharpness-aware minimization.
\newblock In {\em CVPR}, pages 12360--12370, 2022.

\bibitem{liu2022nonuniform}
Zechun Liu, Kwang-Ting Cheng, Dong Huang, Eric~P Xing, and Zhiqiang Shen.
\newblock Nonuniform-to-uniform quantization: Towards accurate quantization via
  generalized straight-through estimation.
\newblock In {\em CVPR}, pages 4942--4952, 2022.

\bibitem{liu2021adam}
Zechun Liu, Zhiqiang Shen, Shichao Li, Koen Helwegen, Dong Huang, and
  Kwang{-}Ting Cheng.
\newblock How do adam and training strategies help bnns optimization.
\newblock In {\em ICML}, volume 139, pages 6936--6946, 2021.

\bibitem{liu2021post}
Zhenhua Liu, Yunhe Wang, Kai Han, Wei Zhang, Siwei Ma, and Wen Gao.
\newblock Post-training quantization for vision transformer.
\newblock {\em NeurIPS}, 34:28092--28103, 2021.

\bibitem{liu2018bi}
Zechun Liu, Baoyuan Wu, Wenhan Luo, Xin Yang, Wei Liu, and Kwang-Ting Cheng.
\newblock Bi-real net: Enhancing the performance of 1-bit cnns with improved
  representational capability and advanced training algorithm.
\newblock In {\em ECCV}, pages 722--737, 2018.

\bibitem{loshchilov2016sgdr}
Ilya Loshchilov and Frank Hutter.
\newblock {SGDR:} stochastic gradient descent with warm restarts.
\newblock In {\em ICLR}, 2017.

\bibitem{loshchilov2018decoupled}
Ilya Loshchilov and Frank Hutter.
\newblock Decoupled weight decay regularization.
\newblock In {\em ICLR}, 2019.

\bibitem{McCann2017LearnedIT}
Bryan McCann, James Bradbury, Caiming Xiong, and Richard Socher.
\newblock Learned in translation: Contextualized word vectors.
\newblock In {\em NeurIPS}, 2017.

\bibitem{mckinstry2019discovering}
Jeffrey~L McKinstry, Steven~K Esser, Rathinakumar Appuswamy, Deepika Bablani,
  John~V Arthur, Izzet~B Yildiz, and Dharmendra~S Modha.
\newblock Discovering low-precision networks close to full-precision networks
  for efficient inference.
\newblock In {\em NIPSW}, pages 6--9. IEEE, 2019.

\bibitem{moosavi2019robustness}
Seyed-Mohsen Moosavi-Dezfooli, Alhussein Fawzi, Jonathan Uesato, and Pascal
  Frossard.
\newblock Robustness via curvature regularization, and vice versa.
\newblock In {\em CVPR}, pages 9078--9086, 2019.

\bibitem{na2022train}
Clara Na, Sanket~Vaibhav Mehta, and Emma Strubell.
\newblock Train flat, then compress: Sharpness-aware minimization learns more
  compressible models.
\newblock {\em arXiv preprint arXiv:2205.12694}, 2022.

\bibitem{nagel2022overcoming}
Markus Nagel, Marios Fournarakis, Yelysei Bondarenko, and Tijmen Blankevoort.
\newblock Overcoming oscillations in quantization-aware training.
\newblock In {\em ICML}, volume 162, pages 16318--16330, 2022.

\bibitem{nilsback2008automated}
Maria-Elena Nilsback and Andrew Zisserman.
\newblock Automated flower classification over a large number of classes.
\newblock In {\em ICVGIP}, pages 722--729. IEEE, 2008.

\bibitem{park2020profit}
Eunhyeok Park and Sungjoo Yoo.
\newblock Profit: A novel training method for sub-4-bit mobilenet models.
\newblock In {\em ECCV}, pages 430--446. Springer, 2020.

\bibitem{parkhi2012cats}
Omkar~M Parkhi, Andrea Vedaldi, Andrew Zisserman, and CV Jawahar.
\newblock Cats and dogs.
\newblock In {\em CVPR}, pages 3498--3505. IEEE, 2012.

\bibitem{paszke2019pytorch}
Adam Paszke, Sam Gross, Francisco Massa, Adam Lerer, James Bradbury, Gregory
  Chanan, Trevor Killeen, Zeming Lin, Natalia Gimelshein, Luca Antiga, et~al.
\newblock Pytorch: An imperative style, high-performance deep learning library.
\newblock {\em NeurIPS}, 32, 2019.

\bibitem{qin2021bibert}
Haotong Qin, Yifu Ding, Mingyuan Zhang, YAN Qinghua, Aishan Liu, Qingqing Dang,
  Ziwei Liu, and Xianglong Liu.
\newblock Bibert: Accurate fully binarized bert.
\newblock In {\em ICLR}, 2022.

\bibitem{rastegari2016xnor}
Mohammad Rastegari, Vicente Ordonez, Joseph Redmon, and Ali Farhadi.
\newblock Xnor-net: Imagenet classification using binary convolutional neural
  networks.
\newblock In {\em ECCV}, pages 525--542, 2016.

\bibitem{ren2015faster}
Shaoqing Ren, Kaiming He, Ross Girshick, and Jian Sun.
\newblock Faster r-cnn: Towards real-time object detection with region proposal
  networks.
\newblock {\em NeurIPS}, 28, 2015.

\bibitem{sandler2018inverted}
Mark Sandler, Andrew Howard, Menglong Zhu, Andrey Zhmoginov, and Liang-Chieh
  Chen.
\newblock Mobilenetv2: Inverted residuals and linear bottlenecks.
\newblock In {\em CVPR}, pages 4510--4520, 2018.

\bibitem{smith2017bayesian}
Samuel~L. Smith and Quoc~V. Le.
\newblock A bayesian perspective on generalization and stochastic gradient
  descent.
\newblock In {\em ICLR}, 2018.

\bibitem{szegedy2015going}
Christian Szegedy, Wei Liu, Yangqing Jia, Pierre Sermanet, Scott Reed, Dragomir
  Anguelov, Dumitru Erhan, Vincent Vanhoucke, and Andrew Rabinovich.
\newblock Going deeper with convolutions.
\newblock In {\em CVPR}, pages 1--9, 2015.

\bibitem{vaswani2017attention}
Ashish Vaswani, Noam Shazeer, Niki Parmar, Jakob Uszkoreit, Llion Jones,
  Aidan~N Gomez, {\L}ukasz Kaiser, and Illia Polosukhin.
\newblock Attention is all you need.
\newblock {\em NeurIPS}, 30, 2017.

\bibitem{wang2018glue}
Alex Wang, Amanpreet Singh, Julian Michael, Felix Hill, Omer Levy, and
  Samuel~R. Bowman.
\newblock {GLUE}: A multi-task benchmark and analysis platform for natural
  language understanding.
\newblock In {\em ICLR}, 2019.

\bibitem{wang2019haq}
Kuan Wang, Zhijian Liu, Yujun Lin, Ji Lin, and Song Han.
\newblock Haq: Hardware-aware automated quantization with mixed precision.
\newblock In {\em CVPR}, pages 8612--8620, 2019.

\bibitem{wang2022learnable}
Longguang Wang, Xiaoyu Dong, Yingqian Wang, Li Liu, Wei An, and Yulan Guo.
\newblock Learnable lookup table for neural network quantization.
\newblock In {\em CVPR}, pages 12423--12433, 2022.

\bibitem{wu2020adversarial}
Dongxian Wu, Shu-Tao Xia, and Yisen Wang.
\newblock Adversarial weight perturbation helps robust generalization.
\newblock {\em NeurIPS}, 33:2958--2969, 2020.

\bibitem{yamamoto2021learnable}
Kohei Yamamoto.
\newblock Learnable companding quantization for accurate low-bit neural
  networks.
\newblock In {\em CVPR}, pages 5029--5038, 2021.

\bibitem{Yang_2019_CVPR}
Jiwei Yang, Xu Shen, Jun Xing, Xinmei Tian, Houqiang Li, Bing Deng, Jianqiang
  Huang, and Xian-sheng Hua.
\newblock Quantization networks.
\newblock In {\em CVPR}, 2019.

\bibitem{zhang2018lq}
Dongqing Zhang, Jiaolong Yang, Dongqiangzi Ye, and Gang Hua.
\newblock Lq-nets: Learned quantization for highly accurate and compact deep
  neural networks.
\newblock In {\em ECCV}, pages 365--382, 2018.

\bibitem{Zhao2020Linear}
Xiandong Zhao, Ying Wang, Xuyi Cai, Cheng Liu, and Lei Zhang.
\newblock Linear symmetric quantization of neural networks for low-precision
  integer hardware.
\newblock In {\em ICLR}, 2020.

\bibitem{zheng2021regularizing}
Yaowei Zheng, Richong Zhang, and Yongyi Mao.
\newblock Regularizing neural networks via adversarial model perturbation.
\newblock In {\em CVPR}, pages 8156--8165, 2021.

\bibitem{zhou2016dorefa}
Shuchang Zhou, Yuxin Wu, Zekun Ni, Xinyu Zhou, He Wen, and Yuheng Zou.
\newblock Dorefa-net: Training low bitwidth convolutional neural networks with
  low bitwidth gradients.
\newblock {\em arXiv preprint arXiv:1606.06160}, 2016.

\bibitem{zhuang2020training}
Bohan Zhuang, Lingqiao Liu, Mingkui Tan, Chunhua Shen, and Ian Reid.
\newblock Training quantized neural networks with a full-precision auxiliary
  module.
\newblock In {\em CVPR}, pages 1488--1497, 2020.

\bibitem{zhuang2018towards}
Bohan Zhuang, Chunhua Shen, Mingkui Tan, Lingqiao Liu, and Ian Reid.
\newblock Towards effective low-bitwidth convolutional neural networks.
\newblock In {\em CVPR}, pages 7920--7928, 2018.

\bibitem{zhuang2021surrogate}
Juntang Zhuang, Boqing Gong, Liangzhe Yuan, Yin Cui, Hartwig Adam, Nicha~C
  Dvornek, James s Duncan, Ting Liu, et~al.
\newblock Surrogate gap minimization improves sharpness-aware training.
\newblock In {\em ICLR}, 2022.

\end{thebibliography}
}

\onecolumn

\begin{center}
	{
		\Large{\textbf{Appendix}}
	}
\end{center}

\renewcommand\thesection{\Alph{section}}
\renewcommand\thefigure{\Alph{figure}}
\renewcommand\thetable{\Alph{table}}
\renewcommand{\theequation}{\Alph{equation}}

\setcounter{table}{0}
\setcounter{figure}{0}
\setcounter{section}{0}

\section{More Implementation Details}
\label{sec:more_implementation_details}
In this section, we provide more implementation details of \methodshortname. 
\revise{As suggested by SAM~\cite{foret2021sharpnessaware} and GSAM~\cite{zhuang2021surrogate}, we apply $m$-sharpness strategy with $m=128$.} For both CNNs and ViTs, we use inception-style pre-processing~\cite{szegedy2015going} without strong data augmentation. Specifically, we randomly crop $224 \times 224$ patches from an image or its horizontal flip counterpart for training. At test time, a $224 \times 224$ centered crop is chosen.
For the hyper-parameter $\rho$, we conduct grid search over \{0.02, 0.05, 0.1, 0.15, 0.2, $\dots$, 1.0\} to find appropriate values \revise{following the common practice in SAM~\cite{foret2021sharpnessaware} and GSAM~\cite{zhuang2021surrogate}}. We put the detailed settings of $\rho$ in Table~\ref{table:rho_details}. \revise{For MobileNetV2, we fine-tune the quantized model with additional learnable layer-wise offsets for activations and knowledge distillation following~\cite{park2020profit,liu2022nonuniform}. As suggested by~\cite{park2020profit,nagel2022overcoming}, we apply BN re-estimation for 10 iterations after training.}
To compute the largest eigenvalue $\lambda_{max}$ of the Hessian of different quantized models on ImageNet, we use the power iteration algorithm following~\cite{dong2019hawq}. To reduce the computational cost, we randomly sample 10k training images for computation. 

\revise{Unless otherwise specified, we do not apply advanced techniques such as knowledge distillation~\cite{zhuang2018towards,park2020profit,zhuang2020training,liu2022nonuniform}, non-uniform quantization~\cite{liu2022nonuniform,Li2020Additive,yamamoto2021learnable}, asymmetric quantization~\cite{bhalgat2020lsq+,park2020profit,liu2022nonuniform}, pre-activation~\cite{Esser2020LEARNED,yamamoto2021learnable,liu2022nonuniform}, weight regularization~\cite{liu2022nonuniform,han2021improving,nagel2022overcoming}, gradient scaling~\cite{lee2021network}, progressive quantization~\cite{zhuang2018towards,park2020profit}, batch normalization re-estimation~\cite{park2020profit,nagel2022overcoming} and iterative training with weight freezing~\cite{park2020profit,nagel2022overcoming}, which can further improve the performance of the quantized models.}

For the transfer learning experiments in Section~5.2, we train all models for 100 epochs. We use SGD with a momentum term of 0.9 for optimization. The learning rate is initialized to 0.01 and decreased to 0 using the cosine annealing. The mini-batch size and the weight decay are set to 64 and 0, respectively.

\begin{table*}[!ht]
\caption{Hyper-parameter $\rho$ for different quantized models on ImageNet.
}
\centering
\scalebox{0.8}
{
\begin{tabular}{cccccccccccc}
\toprule
Network & \multicolumn{2}{c}{ResNet-18} & \multicolumn{2}{c}{ResNet-34} & \multicolumn{2}{c}{ResNet-50} & MobileNetV2 & ViT-S/32 & ViT-S/16 & ViT-B/32 & ViT-B/16 \\
\midrule
Bitwidth & 2 & 4 & 2 & 4 & 2 & 4 & 4 & 4 & 4 & 4 & 4 \\
$\rho$ & 0.30 & 0.65 & 0.65 & 0.90 & 0.65 & 0.95 & 0.40 & 0.01 & 0.01 & 0.01 & 0.01 \\
\bottomrule
\end{tabular}
}
\label{table:rho_details}
\end{table*}

\section{Sensitivity of Hyperparameter $\rho$}
\begin{figure}[!ht]
    \centering
    \includegraphics[height=1.6in]{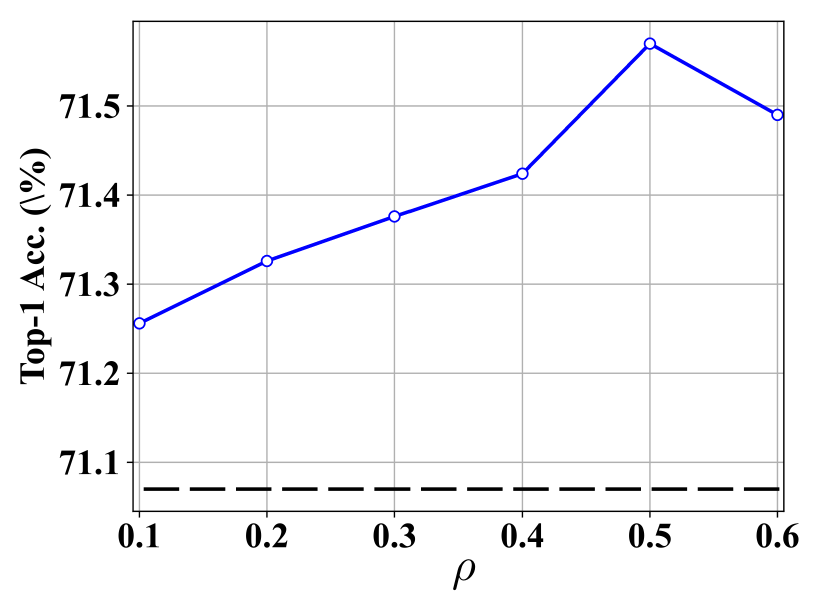}
    \caption{\revise{Performance comparisons with different hyperparameter $\rho$. We apply \methodshortname to obtain 4-bit ResNet-18 on ImageNet. The black dashed line denotes the results of the quantized model trained by SGD.}}
    \label{fig:different_rho}
    \vspace{-0.1in}
\end{figure}

To investigate the sensitivity of hyperparameter $\rho$, we apply SAQ to train 4-bit ResNet-18 with different $\rho$ and show the results in Figure~\ref{fig:different_rho}. From the results, SAQ is relatively insensitive to $\rho$ as it outperforms SGD for a wide range of values. With the increase of $\rho$, the performance of the quantized models improves initially but then deteriorates. On the one hand, increasing perturbation strength helps to improve the generalization performance of the quantized model. On the other hand, too large $\rho$ may incur optimization difficulty and thus lead to sub-optimal performance.

\section{Visualization of the Loss Landscapes}
\label{sec:visualizaiton}
\begin{figure*}[t]
     \centering
     \begin{subfigure}[t]{0.35\textwidth}
         \centering
         \includegraphics[width=0.88\textwidth]{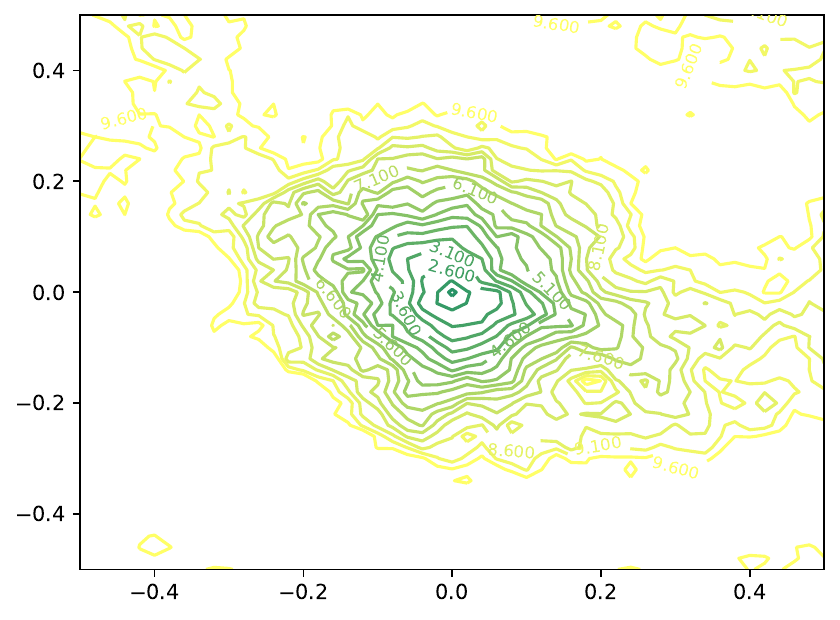}
         \caption{2-bit ResNet-18 obtained by SGD}
         \label{fig:r18_sgd_w2}
     \end{subfigure}
     \hspace{0.2in}
     \begin{subfigure}[t]{0.35\textwidth}
         \centering
         \includegraphics[width=0.88\textwidth]{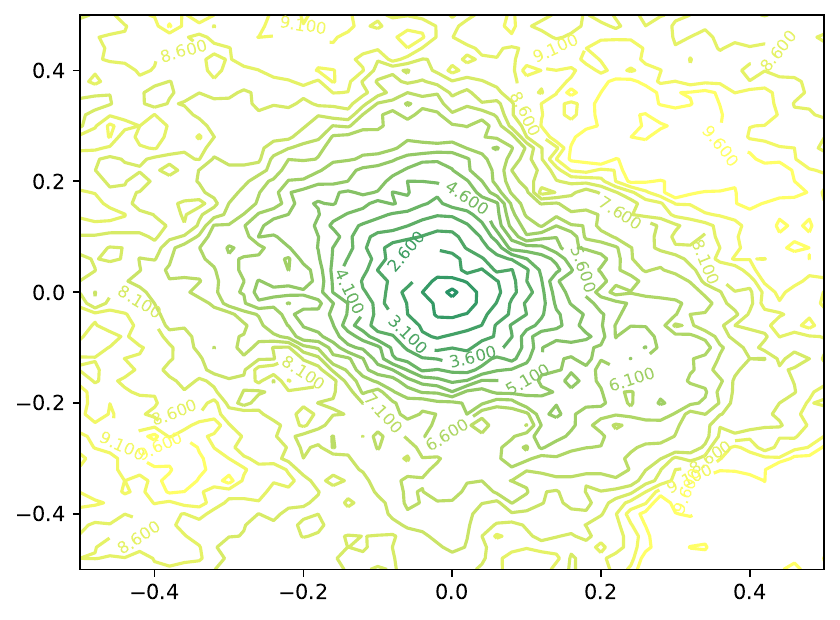}
         \caption{2-bit ResNet-18 obtained by \methodshortname}
         \label{fig:r18_saq_w2}
     \end{subfigure}
     \begin{subfigure}[t]{0.35\textwidth}
         \centering
         \includegraphics[width=0.9\textwidth]{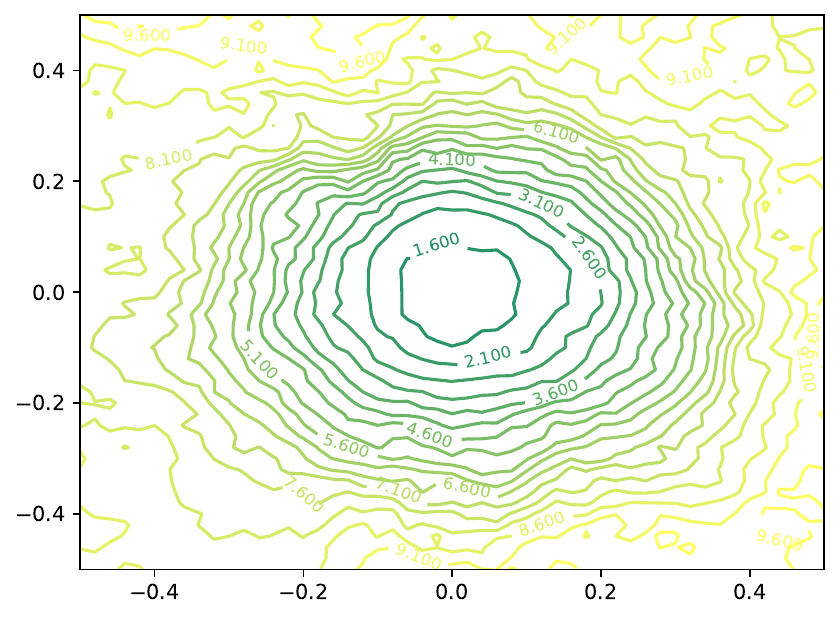}
         \caption{4-bit ResNet-18 obtained by SGD}
         \label{fig:r18_sgd_w4}
     \end{subfigure}
     \hspace{0.2in}
     \begin{subfigure}[t]{0.35\textwidth}
         \centering
         \includegraphics[width=0.9\textwidth]{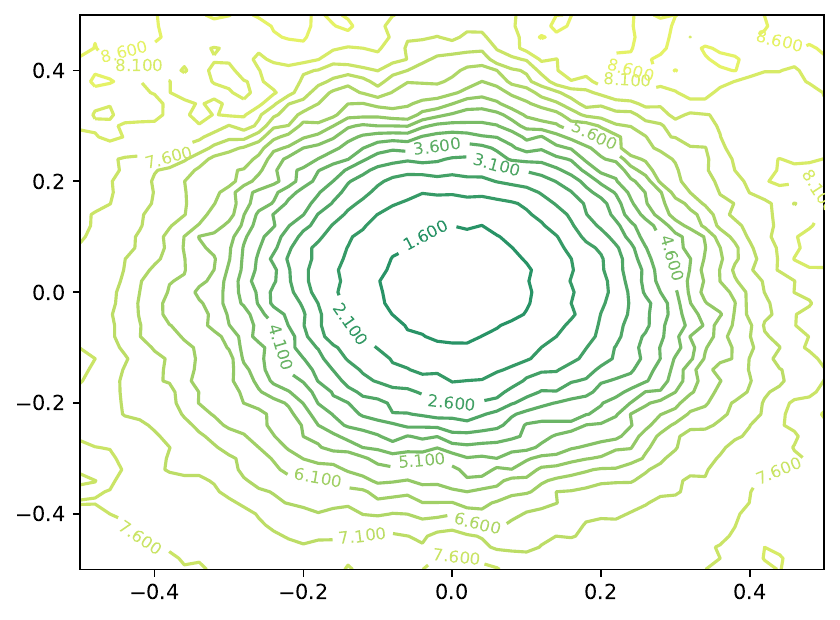}
         \caption{4-bit ResNet-18 obtained by \methodshortname}
         \label{fig:r18_saq_w4}
     \end{subfigure}
        \caption{The loss landscapes of the 2/4-bit ResNet-18 obtained by different methods on ImageNet. }
        \label{fig:r18_loss_landscape}
\end{figure*}

\begin{figure*}[t]
     \centering
     \begin{subfigure}[t]{0.35\textwidth}
         \centering
         \includegraphics[width=0.9\textwidth]{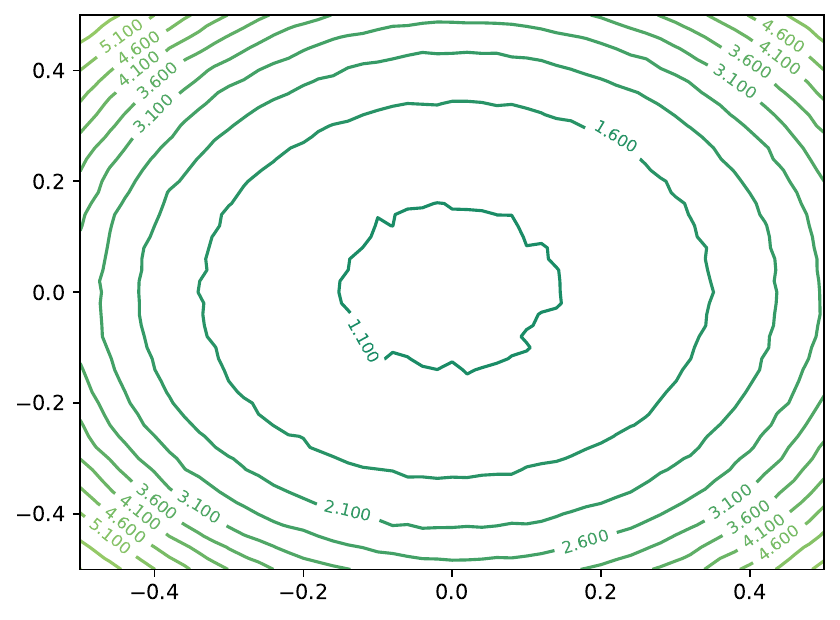}
         \caption{4-bit ViT-B/32 obtained by SGD}
         \label{fig:r50_sgd_w4}
     \end{subfigure}
     \hspace{0.2in}
     \begin{subfigure}[t]{0.35\textwidth}
         \centering
         \includegraphics[width=0.9\textwidth]{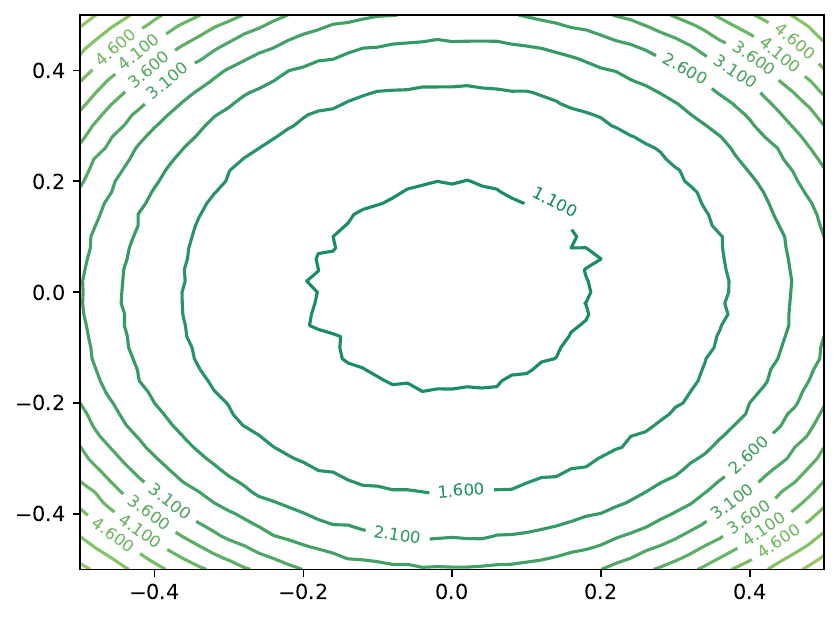}
         \caption{4-bit ViT-B/32 obtained by \methodshortname}
         \label{fig:r50_saq_w4}
     \end{subfigure}
        \caption{The loss landscapes of the 4-bit ViT-B/32 obtained by different methods on ImageNet. }
        \vspace{-0.2in}
        \label{fig:vitb_loss_landscape}
\end{figure*}

In this section, we show the loss landscape of different quantized models on ImageNet using the visualization method in~\cite{Li2018VisualizingTL}. We show the results in Figures~\ref{fig:r18_loss_landscape} and~\ref{fig:vitb_loss_landscape}. The $x$- and $y$-axes of the figures represent two randomly sampled orthogonal directions. From the results, the loss landscapes of the quantized models become smoother and flatter with the increase of bitwidth, 
\jing{suggesting} that smoothing the loss landscapes of the 4-bit quantized models is easier than the 2-bit counterparts. 
Moreover, the loss landscapes of the quantized models obtained by \methodshortname are less chaotic and show larger contour interval compared with the SGD counterpart, indicating that \methodshortname is able to find flatter and smoother minima over SGD.

\clearpage

\end{document}